\documentclass[runningheads,orivec]{llncs}
\usepackage[T1]{fontenc}
\usepackage{graphicx}
\usepackage{booktabs}
\usepackage[misc]{ifsym}
\newcommand{\corr}{(\Letter)}
\usepackage{url}
\usepackage[hidelinks]{hyperref}
\usepackage{algorithm}
\usepackage{subfigure}

\usepackage{bm}
\usepackage{algpseudocode}
\usepackage{amsfonts,amsmath}
\usepackage{dsfont}
\usepackage[table,xcdraw]{xcolor}
\usepackage[capitalise]{cleveref}

\Crefname{section}{Sec.}{Secs.}
\Crefname{figure}{Fig.}{Figs.}
\Crefname{table}{Tab.}{Tabs.}
\Crefname{equation}{Eq.}{Eqs.}
\definecolor{myblue}{HTML}{1f77b4}
\definecolor{mygreen}{HTML}{2ca02c}
\definecolor{myorange}{HTML}{ff7f0e}
\definecolor{darkred}{rgb}{0.55, 0.0, 0.0}
\definecolor{darkblue}{rgb}{0.0, 0.0, 0.55}
\definecolor{darkgreen}{rgb}{0.0, 0.2, 0.13}
\definecolor{mediumorchid}{rgb}{0.73, 0.33, 0.83}
\newcommand{\mat}[1]{\mathbf{#1}}

\begin{document}

\title{Merging Embedded Topics with Optimal Transport for Online Topic Modeling on~Data~Streams}

\titlerunning{StreamETM}
% If the full title of your paper is short enough to also fit in the running head, you can omit the abbreviated paper title here. You can check as follows: if you comment out the \titlerunning line, something will appear in the header of all odd-numbered pages of your PDF from page 3 onward. This something is either the full title (in which case all is well), or the error message "Title Suppressed Due to Excessive Length". If this error message appears, you're going to want to provide an abbreviated title within the \titlerunning command, because if you won't do it, Springer will do it for you.

%N.B.: Author information (both in the \author{} and \authorrunning{} command) should only be present in the Camera-Ready Version of your paper. The version that you initially submit for review, ought to be double-blind. So, when initially submitting your paper, use:
%\author{Author information scrubbed for double-blind reviewing}
\author{Federica Granese\inst{1,3} \corr \and
Benjamin Navet\inst{4} \and\\
Serena Villata\inst{1} \and Charles Bouveyron\inst{2}}
% You may leave out the orcidID information, if you want to.
% Use \corr to indicate the corresponding author. Note the spacing around the \corr command. Only one author can be the corresponding author.

%N.B.: comment out the \authorrunning{} command for the double-blind version of your paper submitted for review. Later, if your paper is accepted, use the command for the Camera-Ready Version.
\authorrunning{F.Granese et al.}
% If there are two authors, write 'A.L. Benjamin and C.C. Broadus Jr.'
% If there are more than two authors, '[...] et al.' is used.

\institute{
Université Côte d’Azur, CNRS, Inria, I3S, Marianne, France
\and Université Côte d’Azur, Inria, CNRS, LJAD, Maasai, France
\and Inria, Defence \& Security mission, France
\and Université Côte d’Azur, 3IA TechPool, France \\
\email{\{firstname.lastname\}@inria.fr}}

\maketitle              % typeset the header of the contribution

\begin{abstract}
Topic modeling is a key component in unsupervised learning, employed to identify topics within a corpus of textual data. The rapid growth of social media generates an ever-growing volume of textual data daily, making online topic modeling methods essential for managing these data streams that continuously arrive over time. This paper introduces a novel approach to online topic modeling named StreamETM. This approach builds on the Embedded Topic Model (ETM) to handle data streams by merging models learned on consecutive partial document batches using unbalanced optimal transport. Additionally, an online change point detection algorithm is employed to identify shifts in topics over time, enabling the identification of significant changes in the dynamics of text streams. Numerical experiments on simulated and real-world data show StreamETM outperforming competitors. We provide the code publicly available at \url{https://github.com/fgranese/StreamETM}.

\end{abstract}

\section{Introduction}
With the rapid expansion of social media and digital communication, vast amounts of textual data are continuously generated and distributed across various platforms. This growing volume of information necessitates automated methods for efficient information retrieval. 
In this context, topic models are powerful statistical tools for uncovering the hidden semantic structure within a collection of documents~\cite{dieng2020topic}. Specifically, these models aim to identify latent topics based on word co-occurrence patterns. Each topic represents a coherent semantic concept and is characterized by a group of related words. For instance, a topic related to \texttt{sports} may include words such as \texttt{baseball}, \texttt{basketball}, and \texttt{football}~\cite{wu2024survey}.  Topic models have been widely applied to analyze various types of textual data, including fiction, non-fiction, scientific publications, and political texts~\cite{boyd2017applications}.
However, most of the existing work on topic modeling focuses on \textit{offline} settings, where the model is trained on a fixed dataset (batch) and remains static. However, with the continuous generation of new content, there is a growing need for models that can operate in an \textit{online} setting. Typical examples are news agencies that release their clients' news in real time or social networks that continuously deliver their users' posts to the network. In these scenarios, topic modeling algorithms must continuously update as new documents arrive.

Recent solutions for online topic modeling are often built upon BERTopic~\cite{grootendorst2022bertopic}, a model that generates topic representations in three main steps: document embeddings, dimensionality reduction, and clustering. Unfortunately, since these models rely on pre-trained language models, fine-tuning the parameters for each step as new data arrives is challenging. This can make maintaining an efficient and adaptive process difficult as the data evolves in real-time.

In online settings, another challenge is to automatically associate new topics with existing ones. Indeed, topics are not static, but they evolve over time, often shifting in meaning and representation. Nevertheless, most existing models rely on static clustering methods or fixed word distributions, making it difficult to track these changes effectively. For example, before 2022, discussions on \texttt{AI} were likely dominated by terms like \texttt{transformers} and \texttt{GAN}, whereas today, they focus more on \texttt{LLM}.

Finally, users of these online methods must be able to detect significant shifts in the model's dynamics. Indeed, analysts monitoring data flows of this type are particularly interested in being alerted when a sudden or significant change has occurred in the data flow. In this case, the user can analyze the changes between topics and take appropriate decisions and actions. To our knowledge, this feature is not currently offered as part of online topic modeling methods.

This work addresses these limitations through the following contributions:
\begin{enumerate}
    \item \textbf{We explore the potential of optimal transport for topic association and discovery}, demonstrating its effectiveness in aligning evolving topics and ensuring coherent topic transitions over time, and its superiority over Euclidean or Cosine similarities for this task.
    \item \textbf{We introduce StreamETM, an online version of the Embedded Topic Model (ETM)}. StreamETM combines a variational inference strategy for the ETM model, applied sequentially on consecutive time windows, with a merging approach based on unbalanced optimal transport. 
    \item \textbf{We complete StreamETM with a change point detection algorithm}, allowing the automatic determination of significant changes in the dynamics of the studied documents. To our knowledge, StreamETM is the only unsupervised and online approach proposed for the complex tasks of online topic modeling and change point detection on text data streams.
\end{enumerate}

\subsection{Related works}
\label{sec:related}
\paragraph{Offline setting.}
Topic modeling was initially developed using heuristic approaches and was later studied with a statistical perspective two decades ago. The Latent Semantic Index~\cite{deerwester1990indexing}(LSI) is considered the first work to provide statistical foundations for this task. Building on this, Probabilistic LSI (pLSI)~\cite{hofmann1999probabilistic} introduced a mixture model, where each component represents a specific topic and defines a corresponding vocabulary distribution. 
However, LSI lacks a generative model at the document level and is prone to overfitting.
In 2003, Blei et al. proposed Latent Dirichlet Allocation (LDA)~\cite{blei2003latent}, which models topic proportions using a Dirichlet distribution. Extensions include deep generative models, such as~\cite{srivastava2017autoencoding}, which introduced a variational distribution parametrized by a neural network and Wasserstein autoencoders~\cite{nan2019topic}. A successive evolution of the LDA is with the Embedded Topic Model~\cite{dieng2020topic} (ETM), which allowed using deep embeddings to represent both the words and the topics in the same vector space. 
Specifically, these embeddings are
part of the decoder and can be pre-trained on large datasets to incorporate semantic meaning. 
More recently, language models such as BERT~\cite{devlin2019bert} have been used for topic modeling. BERTopic~\cite{grootendorst2022bertopic} is a topic model that generates topic representations in three steps: first, each document is embedded using a pre-trained language model; second, UMAP reduces the embeddings' dimensionality for optimized clustering with HDBSCAN; finally, topic representations are extracted from the clusters using a custom class-based TF-IDF (c-TF-IDF) variation. 
Finally, recently, topic modeling has been explored by \underline{prompting} large language models to generate a set of
topics given an input dataset~\cite{pham2023topicgpt,doi2024topic}.

\paragraph{Online setting.}
On the one hand, LDA was first extended to an online version in~\cite{hoffman2010online} with a stochastic optimization algorithm using a natural gradient step to optimize the variational Bayes lower bound as data arrives. However, this approach is not suited for streams of documents that cannot be stored. The solution in~\cite{amoualian2016streaming} addresses this by extending LDA to document batches using copulas.
On the other hand, two versions of BERTopic can be used in an online setting, namely MergeBERT\footnote{\url{https://maartengr.github.io/BERTopic/getting_started/merge/merge.html}} and OnlineBERT\footnote{\url{https://maartengr.github.io/BERTopic/getting_started/online/online.html}}.
MergeBERT is a pseudo-online variant of the original BERTopic model~\cite{grootendorst2022bertopic}. Topic models are merged sequentially by comparing their topic embeddings. If topics from different models are sufficiently similar (w.r.t. cosine similarity), they are considered the same, and the topic from the first model in the sequence is retained. However, if topics are dissimilar, the topic from the latter model is added to the former set. Crucially, this process does not involve an actual merging of topic representations, meaning that the words associated with a topic do not evolve over time as new models (and, therefore, new documents) are introduced. 
OnlineBERT preserves the embedding transformation of the documents and the final c-TF-IDF approach used in BERTopic while introducing an online variant for the dimensionality reduction step (IncrementalPCA), a MiniBatchKMeans for clustering, and an online CountVectorizer for tokenization to update out-of-vocabulary words and to prevent the sparse bag-of-words matrix from growing excessively large. Compared to MergeBERT, OnlineBERT, while being truly online, loses some of the advantages of the former model. For instance, UMAP is generally better at preserving complex, non-linear relationships, which can lead to more coherent topics. Furthermore, the combination of IncrementalPCA and MiniBatchKMeans may result in the over-proliferation of subtopics over time. A single topic could be split into multiple subtopics as new data arrives, often leading to unnecessary topics that are difficult to interpret. We emphasize that the online and dynamic settings for topic modeling are fundamentally distinct. In online topic models, data is processed incrementally, with topics being updated in real time as new documents arrive. In contrast, dynamic topic modeling works \textit{a posteriori} and captures the temporal evolution of topics by analyzing them across fixed time intervals (e.g., weeks or years).

\paragraph{Optimal transport and topic modeling.} Optimal transport has already been used in a few situations related to topic modeling. We refer to~\cite{dhanania2024interactive}, which employs optimal transport for label name-supervised topic modeling, assigning documents to predefined topics based on semantic similarities computed from pre-trained LMs/LLMs. Similarly, in~\cite{wu2024fastopic}, documents are embedded into an $H$-dimensional semantic space using a pre-trained transformer model, such as BERT. In this approach, topics and words are randomly projected into the same semantic space, with their embeddings jointly optimized alongside the transport maps. Our work differs in several key aspects. We leverage optimal transport to merge topic embeddings rather than to establish document-topic associations. Consequently, the transport map is applied to objects within the same semantic space. Moreover, unlike prior work, our approach does not employ the transport map during training to optimize embeddings. Instead, as discussed in~\cref{sec:discussion}, it can also be utilized at evaluation time to align predicted and ground-truth topics. Lastly, our framework operates in an online setting, with data coming over time. 

\section{Preliminaries on the Embedded Topic Model}
\label{sec:preliminaries}
%\paragraph{The Embedded Topic Model.}
Let us consider  for the moment a corpus $\mathcal{W} = \{\mat{W}^{(1)}, \dots, \mat{W}^{(D)}\}$ of $D$ documents, where the vocabulary consists of $V$ distinct words. Each $\mat{W}^{(d)}$ contains $N_d$ words and is represented as
$\mat{W}^{(d)} = (\mat{w}^{(d)}_1, \dots, \mat{w}^{(d)}_{N_d})\in\{0, 1\}^{N_d\times V}$
where each word $\mat{w}^{(d)}_j$ is a one-hot vector, meaning that $w^{(d)}_{ji} = 1$, if the  $j$-th word in document $d$ is the $i$-th word in the vocabulary, 0, otherwise. 

A ``topic'' is represented as a full distribution over the vocabulary, and a document is assumed to come from a mixture of topics, where the topics are shared across the corpus, and the mixture proportions are unique to each document. Specifically, a \textit{topic} $k$ is represented by a vector $\beta_k \in \Delta_V$, where $\Delta_V$ is the $V$-dimensional simplex. We denote the \textit{topic matrix} as $\bm{\beta} = (\beta_1, \dots, \beta_K)\in\mathbb{R}^{V\times K}$.
In Embedded Topic Models~\cite{dieng2020topic} (ETM), both words and topics are represented using embeddings. ETM first embeds the vocabulary in an $L$-dimensional space and represents each document in terms of $K$ \textit{latent topics}. We call the embeddings of the words $\bm{\rho}=(\rho_1, \dots, \rho_V)\in\mathbb{R}^{L\times V}$ providing the representation of the words in a $L$-dimensional space. Similarly, a \textit{latent topic} $k$ is represented by a vector $\alpha_k\in\mathbb{R}^L$ and we denote the \textit{latent topic matrix} $\bm{\alpha} = (\alpha_1, \dots, \alpha_K)\in\mathbb{R}^{L\times K}$.
In this context, the topic distribution over the vocabulary is assumed to be 
% \begin{align}
    % \label{eq:softmax}
$\beta_k = \text{softmax}(\bm{\rho}^T \alpha_k)$,
% \end{align}
where the ETM assigns a high probability to word $j$ in topic $k$ by measuring the agreement between the word embedding and the topic embedding.
We refer to the seminal paper of Dieng et al.~\cite{dieng2020topic} for the full description of the generative process of the $d$-th document under the ETM. Overall, the ETM model assumes that each document $d$ is sampled from a mixture of topics with its proportion denoted as $\theta_d=\text{softmax}(\delta_d)$ where $\delta_d\sim\mathcal{N}(0, I)$. For each word $j$ in the document, a topic assignment $z^{(d)}_{j}$ is sampled from a categorical distribution parameterized by $\theta_d$. The word $\mat{w}^{(d)}_{j}$ is then generated via a softmax transformation of the inner product between the word and topic embeddings. 

ETM employs variational inference to approximate the intractable likelihood of observing $\mathcal{W}$ given $\bm{\alpha}$ and $\bm{\rho}$, using a mean-field assumption where the variational distribution $q_{\phi}$ factorizes over documents. A variational autoencoder (VAE) models this distribution as a Gaussian with parameters learned by a neural network. The Evidence Lower Bound (ELBO) 
$$\mathcal{L}(\mathcal{W}, \bm{\alpha}, \bm{\rho}; q_{\phi}) = \, \mathbb{E}_{q_{\phi}}\left [\log p(\mathcal{W}, \delta \, |\, \bm{\alpha}, \bm{\rho}\right] - \mathbb{E}_{q_{\phi}} \left [\log q_{\phi}(\delta) \right],
$$ is optimized via the reparameterization trick and stochastic gradient descent.

 \section{The Stream Embedded Topic Model}
\label{sec:stream_etm}
Let us consider a stream of documents arriving as batches at discrete time steps, $\mathcal{W}_{[1:T]} = \{ \mathcal{W}^{(1)}, \dots, \mathcal{W}^{(t-1)}, \mathcal{W}^{(t)}, \mathcal{W}^{(t+1)} , \dots, \mathcal{W}^{(T)}\}$, where each $\mathcal{W}^{(i)}$ in $\mathcal{W}_{[1:T]}$ represents a corpus of documents as defined in~\cref{sec:preliminaries}. 

\subsection{Learning an ETM model on the current batch $\mathcal{W}^{(t)}$}
At each time step, we aim to learn a new ETM model that, based only on the corpus of documents available at the current time step and the latent topic embeddings from the previous step, can accurately link past topics to present ones while also identifying new topics. This scenario differs from a dynamic system, where it is assumed that all information is available at the final time step $T$. In contrast, our setting operates online, where only the data observed up to the current time step can be used for learning. The model must continuously adapt without access to future observations, as in real-time applications. 

We will refer with $M_{\mathcal{W}^{(t-1)}, \bm{\alpha}^{(t-1)},\rho}\equiv M^{(t-1)}$ to the ETM model at time step $t-1$ where $\bm{\alpha}^{(t-1)}$ is the latent topic matrix at time $t-1$, we assume the embeddings of the words $\bm{\rho}$ to be constant over time. Similarly, we will denote the topic matrix at time $t-1$ as  $\bm{\beta}^{(t-1)}$.
At time $t$, our goal is first to maximize 
\begin{align}
\label{eq:elbo_t}
    \mathcal{L}(\mathcal{W}^{(t)}, \Tilde{\bm{\alpha}}^{(t)}, \bm{\rho}; q_{\phi^{(t)}}) = \, \mathbb{E}_{q_{\phi^{(t)}}}\left [\log p(\mathcal{W}^{(t)}, \delta^{(t)} \, |\, \Tilde{\bm{\alpha}}^{(t)}, \bm{\rho}\right] - \mathbb{E}_{q_{\phi^{(t)}}} \left [\log q_{\phi^{(t)}}(\delta^{(t)}) \right],
\end{align}
following the classical offline ETM models described in~\cref{sec:preliminaries}. Therefore, we seek an appropriate merging strategy\footnote{Note that, at $t=1$, no merging strategy is applied.} $g$
to map the previously learned topic embedding space and the current one into a new representation, and we impose $\bm{\alpha}^{(t)} = g(\Tilde{\bm{\alpha}}^{(t)}, \bm{\alpha}^{(t-1)})$.
Finally, $M^{(t)}$ is obtained by optimizing a second time~\cref{eq:elbo_t} using stochastic gradient descent, with $\bm{\alpha}^{(t)}$ and $\bm{\rho}$ kept fixed.

\subsection{Optimal transport for merging and discovering topics}
\label{sec:optimal_topic}
We now analyze the problem of determining an \textit{effective} strategy $g$ for identifying the topics in $\bm{\alpha}^{(t-1)}$ that are most similar to those in $\Tilde{\bm{\alpha}}^{(t)}$, enabling us to merge these topic embeddings while incorporating the new topics present in $\Tilde{\bm{\alpha}}^{(t)}$.

\subsubsection{Transport map computation.}
We recall that $$\bm{\alpha}^{(t-1)} = (\alpha^{(t-1)}_1, \dots, \alpha^{(t-1)}_K) \in \mathcal{A}^{(t-1)} \subseteq \mathbb{R}^{L\times K}$$ and $$\Tilde{\bm{\alpha}}^{(t)} = (\tilde{\alpha}^{(t)}_1, \dots, \tilde{\alpha}^{(t)}_J) \in \mathcal{A}^{(t)} \subseteq \mathbb{R}^{L\times J},$$ where $J$ can be either different from or equal to $K$.  
Let $a = \frac{1}{K} \sum_{i=1}^{K} \delta_{\alpha^{(t-1)}_i}$ and $\Tilde{a} = \frac{1}{J} \sum_{i=1}^{J} \delta_{\Tilde{\alpha}^{(t)}_i}$ be the two discrete distributions of mass on $\mathcal{A}^{(t-1)}$ and $\mathcal{A}^{(t)}$. We aim to find the least costly way to shift the mass (i.e., the topics) from the previous time step to the current one. To this end, we formulate the problem as Unbalanced Optimal Transport (UOT)~\cite{benamou2003numerical}, a relaxed version of OT where the total mass of each source (the topics of $\Tilde{\bm{\alpha}}^{(t)}$) can be spread across multiple targets (the topics of $\bm{\alpha}^{(t-1)}$):
 \begin{align}
    \mat{T}^{\star} =  \underset{\mat{T}\in\mathbb{R}^{J\times K}_{+}}{\arg\min}\,\langle \mat{C}, \mat{T}\rangle + \lambda_{\Tilde{a}} D_{\psi}\left(\mat{T}\mathds{1}_J, \Tilde{a}\right) + \lambda_{a} D_{\psi}\left(\mat{T}^{\top}\mathds{1}_K, a\right),
\end{align}
where $\langle \cdot, \cdot\rangle$ is the Frobenius inner product, and $D_{\psi}(\cdot, \cdot)$ is the Bregman divergence that penalizes violations of the marginal constraints. Additionally, $\lambda_a\in\mathbb{R}_{+}$ (resp. $\lambda_{\Tilde{a}}\in\mathbb{R}_{+}$) represents the penalty associated with $a$ (resp. $\Tilde{a}$). Moreover, $\mat{C}\in\mathbb{R}^{J\times K}_{+}$ is the cost-matrix in which the entries $C_{jk}$ encode the cost of moving $\Tilde{\alpha}^{(t)}_{j}$ towards $\alpha^{(t-1)}_{k}$. In this particular setting, as we deal with text, we chose the cosine similarity as the cost function. Finally, $\mathds{1}_{(\cdot)}$ represents the vector of dimension $(\cdot) \times 1$, which is used to ensure that the sum per row and column does not diverge significantly from the original distributions $a$ and $\Tilde{a}$. In this way, only a portion of the total mass is transported, and the total mass can be unbalanced between the sources and targets due to the constraint relaxation.
Intuitively, a sparse transport matrix indicates that mass is transferred only between semantically similar topics, while distant topics receive no transport.

Note that the UOT problem can be efficiently recast as a non-negative penalized linear regression problem. We refer to~\cite{chapel2021unbalanced} for additional details. 

\subsubsection{Merging topics and discovery of new ones.} For each $\Tilde{\alpha}^{(t)}_j$, we determine the corresponding target topic by identifying the index where the transport plan assigns the highest mass. Specifically, we select the topic 
$k^\star$ that maximizes the transport matrix entry, given by:
$k^{\star} = \arg\max_{k\in\{1, \dots, K\}}\,T^{\star}_{jk}$. Therefore,
\begin{align}
    \alpha^{(t)}_j = \omega \Tilde{\alpha}^{(t)}_j + (1 - \omega)\alpha^{(t-1)}_{k^\star},
\end{align}
where $\omega\in [0, 1]$ is a memory parameter. Otherwise, if no mass has been transported from $j$, meaning for all $k\in\{1, \dots, K\}$, $T^\star_{jk} = 0$, the $j$th topic is a new one and can be added to the set of topics: $\alpha^{(t)}_{J+1} = \Tilde{\alpha}^{(t)}_j$.

\subsection{Change point detection}
To monitor the significant changes in the dynamic of the data stream
we analyze, we propose to add a change point detection step to our
approach. In addition to the detection of new topics (topics
that are added in the merged model), we propose to make use of the
online Bayesian changepoint detection (OCPD~\cite{adams2007bayesian})
algorithm to monitor significant changes in the sequence of merged
models $\{M^{(1)},M^{(2)},\dots,M^{(T)}\}$. We propose to apply
the OCPD algorithm to time series of topic distributions over
the documents at different time steps. It is worth highlighting that OCPD is, to this date, the most performant change point detection method able to work in a fully online framework and can issue alerts on the fly.

\section{Experimental Setting}
\label{sec:setting}
We describe the experimental setting for evaluating StreamETM. Designing this setting posed several challenges. First, since we consider an online and unsupervised setting, our evaluation goes beyond assessing model performance at individual time steps; we also analyze the overall interaction dynamics induced by merging topic embeddings. Second, as we are working with topics, relying solely on human judgment for evaluation is insufficient, requiring us to explore alternative quantitative metrics. 
% Appendix A.1 describes the evaluation metrics.
Note that we compare with online (not dynamic) topic models, as dynamic approaches lack incremental learning, which is central to our study (cf.~\cref{sec:related}).
%\Cref{app:evaluation} describes the evaluation metrics.

\subsection{Datasets}
We consider the \texttt{20kNewsGroup}\footnote{\url{http://qwone.com/~jason/20Newsgroups/}} dataset as text corpora for our experiments, comprising around 18k newsgroup posts on 20 topics.
We confine to 5 of the 20 topics and randomly draw 15 times approximately 5k samples from the total datasets\footnote{The same sample may appear in multiple datasets. However, each dataset would have been too small without repetition when partitioned across different time steps.}. We partition the 5k samples into 500 sample batches to simulate a $\approx$10 time steps scenario.
For each time step, the corresponding dataset has been pre-processed by first lemmatizing the text, removing lowercase and punctuation, filtering out stopwords (cf. \texttt{nltk.corpus.stopwords}), and eliminating low-frequency words (words appearing only once) and those appearing in more than 70\% of documents to reduce overly frequent terms. 
The topic distributions are computed considering practical use cases.\\

\noindent\textbf{Our practical use-cases}.
We simulate the online setting by assuming that each time step $\bm\tau^{(i)}$, $i=1, \dots, T$, is represented as a distribution:\\

\noindent a) \textsc{Custom}: A designed setting where the topics are intentionally chosen to be sufficiently distinct. At each time step, at most four out of five topics are \textit{active} (\cref{fig:original_cs}).
Topics: \texttt{autos}, \texttt{sport}, \texttt{medicine}, \texttt{space}, \texttt{religion}.\\

\noindent b) \textsc{Dynamic}: Text corpora with significant temporal shifts in topic relevance. At each time step $i$, the activity of each topic $k$ is determined independently. A binary variable $z^{(i)}_k$ is drawn from a Bernoulli distribution $z^{(i)}_k \sim \text{Bernoulli}(p)$, where $p\in[0, 1]$ represents the probability that a topic remains \textit{active}. For each time step, the unnormalized proportion of topics is: $\tau^{(i)}_k = z^{(i)}_k \cdot \text{Dir}_k(\alpha)$, $\alpha > 1$. Finally, the topic proportions $\bm{\tau}^{(i)}$ are normalized to ensure they sum to 1 (\cref{fig:original_ex}). Topics are randomly chosen: 
\texttt{computer}, \texttt{sale}, \texttt{cryptography}, \texttt{religion}, \texttt{mideast}.\\

\subsection{Architectures and training procedure}
\paragraph{StreamETM.} At each time step, we trained an ETM on English text using fixed GloVe embeddings from the \texttt{glove-wiki-gigaword-300} vocabulary, truncated to the first 15k words. The model was initialized with 3 topics: an 800-dimensional hidden layer for the encoder and 300-dimensional word embeddings.
The topic embeddings were initialized using Xavier uniform initialization at time step 0, while in subsequent iterations, they were set to the values computed at the previous time step, following the strategy described in~\cref{sec:optimal_topic}. The training was performed over 3k epochs with a batch size of 1000, a learning rate of 0.01, and a weight decay of 0.006 using the Adam optimizer.
Regarding the UOT procedure, we use the Cosine Distance for the cost map. The transport map is computed using the Python function \texttt{ot.unbalanced.mm\_unbalanced}, with KL divergence and marginal relaxation at 0.09.

\paragraph{MergeBERT.}
We used the \texttt{paraphrase-multilingual-mpnet-base-v2} model from SentenceTransformers to generate document embeddings. These embeddings were processed using UMAP for dimensionality reduction, with 10 components, a minimum distance of 0.1, and cosine similarity. HDBSCAN was applied for clustering with Euclidean Distance and a minimum cluster size of 3. 
We improved term weighting using the ClassTfidfTransformer with BM25.
BERTopic was used for topic modeling, with the PartOfSpeech model for enhanced text representation. 
The cosine similarity threshold for merging topics was set to 0.7. A lower threshold would lead to over-proliferation of topics, while a higher value could cause the newly formed topics to collapse.

\subsection{Evaluation metrics}
\label{sec:evaluation}
\paragraph{Qualitative.} We analyze the distribution of topics over time and visually compare the original distribution with the predicted ones. In addition, we examine the top five words associated with each topic at each time step. Since each setting involves 15 training runs, we manually align topic indices across executions. On average, MergeBERT identifies more than 20 topics at each time step. Therefore, we focus on topics that are more similar to the targeted ones for this metric.

\paragraph{Quantitative.} We measure topic quality in terms of topic coherence (TC)~\cite{mimno2011optimizing} 
and topic diversity (TD)~\cite{dieng2020topic}. 
Topic coherence is the average pointwise mutual information of two words drawn randomly from the same document: the most likely words in a coherent topic should have high mutual information. In contrast, topic diversity is the percentage of unique words in the top 10 words of all topics. Intuitively, topic diversity measures how varied the overall topics are.

\paragraph{Online change point detection.} We apply an online change point detection (OCPD) algorithm to the predicted topic distributions, using the R package \texttt{ocp} to analyze topic proportions over time and detect significant rupture points. If the predicted distributions closely resemble the original ones, the algorithm should identify rupture points at approximately the same time steps. To avoid the need for manual topic alignment across different training runs, we consider a rupture point to be correctly identified (true positive) if the algorithm detects \textit{any} change at the same time step as in the original distribution or within one-time step before or after, regardless of the specific topic. We compute ROC curves based on different threshold values of the OCPD (between 0 and 1).

\section{Numerical Experiments}
\label{sec:discussion}
This section examines StreamETM from multiple perspectives. We first highlight the advantages of optimal transport for topic merging and discovery, followed by a quantitative evaluation. Finally, we analyze the approach from qualitative, quantitative, and online change-point detection perspectives.

\subsection{Impact of optimal transport for topic merging and discovery}
\subsubsection{Comparison with Euclidean Distance.}
\begin{figure}[ht]
    \centering
    \subfigure[No perturbation applied]{
        \includegraphics[width=0.3\columnwidth]{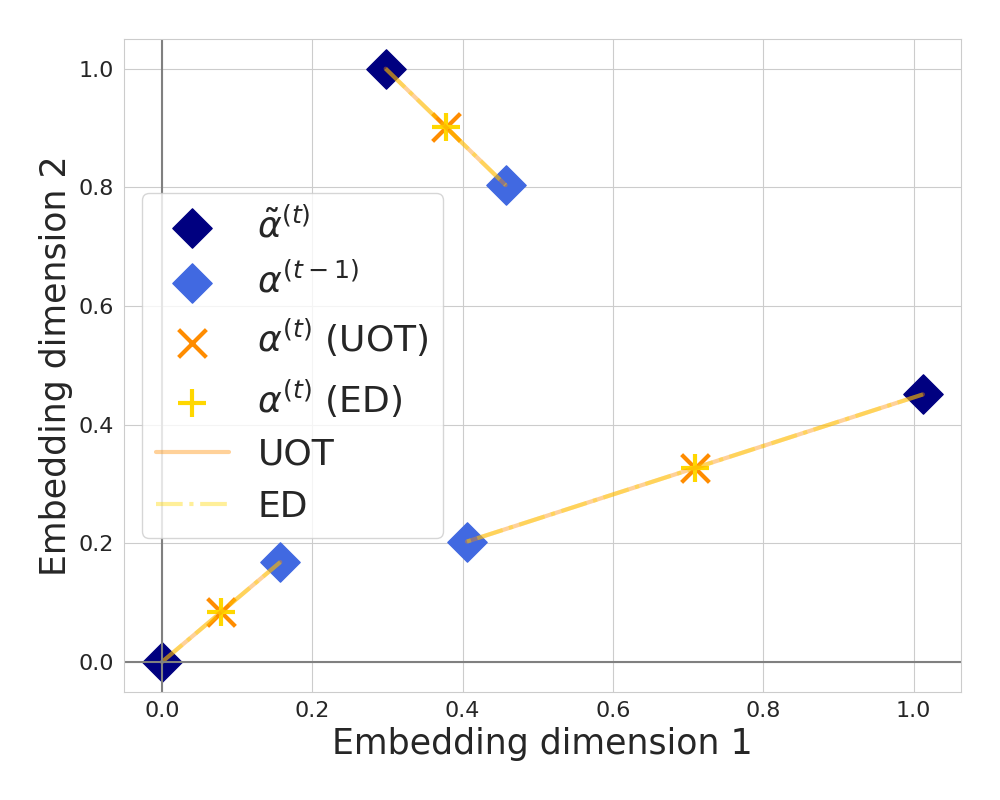}
        \label{fig:no_noise}
    }
    \hfill
    % \hspace{0.5cm}
    \subfigure[Perturbation on a topic]{
        \includegraphics[width=0.3\columnwidth]{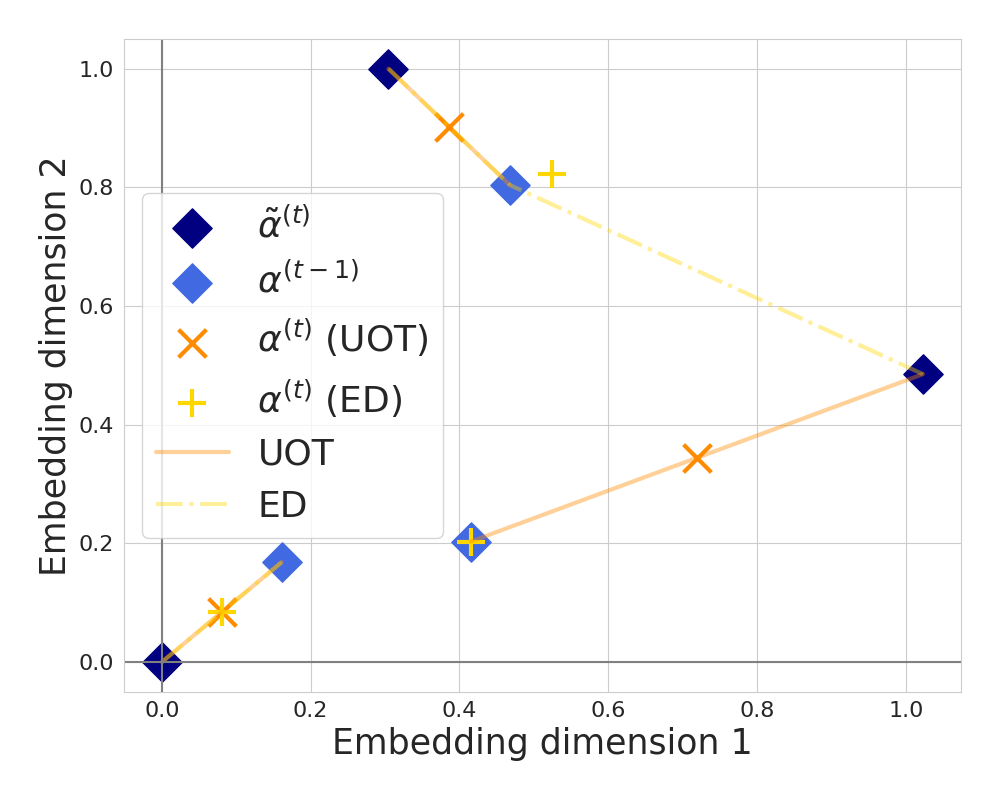}
        \label{fig:noise1}
    }
    \hfill
    \subfigure[New topics]{
        \includegraphics[width=0.3\columnwidth]{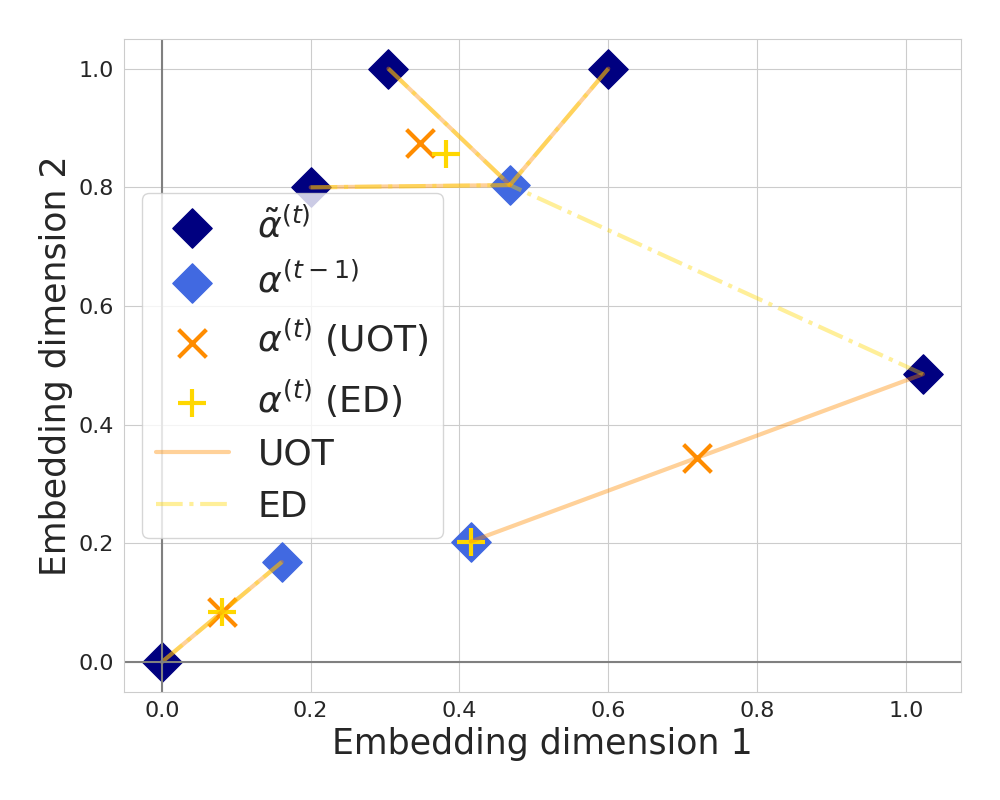}
        \label{fig:noise2}
    }
    \caption{Topic embeddings in a Euclidean space. On the left, the setting is without perturbation, and on the center and the right, a perturbation is added to the dark blue diamond at the position $(1.01,0.45)$. 
    Dark blue diamonds represent topic embeddings at time $t-1$, while light blue markers indicate topic embeddings at time $t$ before merging. The merged embeddings obtained via UOT are shown as `$\times$', whereas those obtained using ED are shown as `$+$'. Dashed lines connect topics matched by UOT, while dot-dashed lines indicate associations based on ED.}
    \label{fig:toy1}
\end{figure}
In~\cref{fig:toy1}, we illustrate the role of unbalanced optimal transport (UOT) in topic merging, comparing it to the classical Euclidean Distance (ED) within a 2D Euclidean space. While ED evaluates pairwise topic distance, disregarding the overall distribution, UOT accounts for the global structure. As a result, \textit{(i)} ED may introduce spurious correlations;
\textit{(ii)} The merge based on ED could be more sensitive to small perturbations of the input.
For simplicity, we model the topics at time $t-1$ (in light blue) as samples from a normal distribution. Similarly, the topics at time $t$ (in dark blue) are generated by perturbing each topic at time $t$ with an additional value drawn from the same normal distribution. To compute the transport map, we use the procedure previously described but consider the Euclidean cost matrix.
Initially, ED and UOT are equally mapping the topics at time $t$ to the ones $t-1$, resulting in generating the same new topics, represented as a `$+$' for ED and a `$\times$' for UOT (cf.~\cref{fig:no_noise}). However, when introducing a small perturbation to the input—specifically shifting the dark blue point initially at coordinates $(1.01, 0.45)$ to 
$(1.02, 0.49)$ in~\cref{fig:noise1}—we observe that while UOT remains stable, ED yields unintended new associations, leading to spurious topics.
Finally, as new topics emerge, as shown in \cref{fig:noise2}, we observe that the newly merged topic is moving closer to the one created with UOT. However, the topic at $(1.02, 0.49)$ is almost completely lost in ED. We can imagine that while these behaviors can be easily crafted in a lower-dimensional space, more complex reactions could arise in a higher-dimensional space, especially when considering the issue of topic overproliferation. This is particularly relevant since most metrics do not account for the global structure of the distributions.
\subsubsection{Comparison with Cosine Distance.}
Similar to our approach with Euclidean Distance, we now demonstrate the role of UOT compared to Cosine Distance (CD). 
We first evaluate UOT and CD to merge the topic embeddings and check the discovery performance in subsequent steps. Specifically,  we consider 7 of the 20 discussion topics in the \texttt{20kNewsGroup}; we randomly draw 1k documents from these topics for two subsequent time steps whose topic distributions are fixed and obtained from a Dirichlet distribution of parameter 1. 
We expect our approach to merge 3 common topics and to detect that 2 new ones should not be merged. 
\Cref{tab:acc_disc} reports the topic merging and discovery accuracies (the closer to 1, the better) averaged on 50 simulated document sets. As can be seen, the UOT approach is globally more efficient than the other approaches. 
\begin{figure}[ht]
    \centering
    \subfigure[Optimal Transport]{
        \includegraphics[width=0.45\columnwidth]{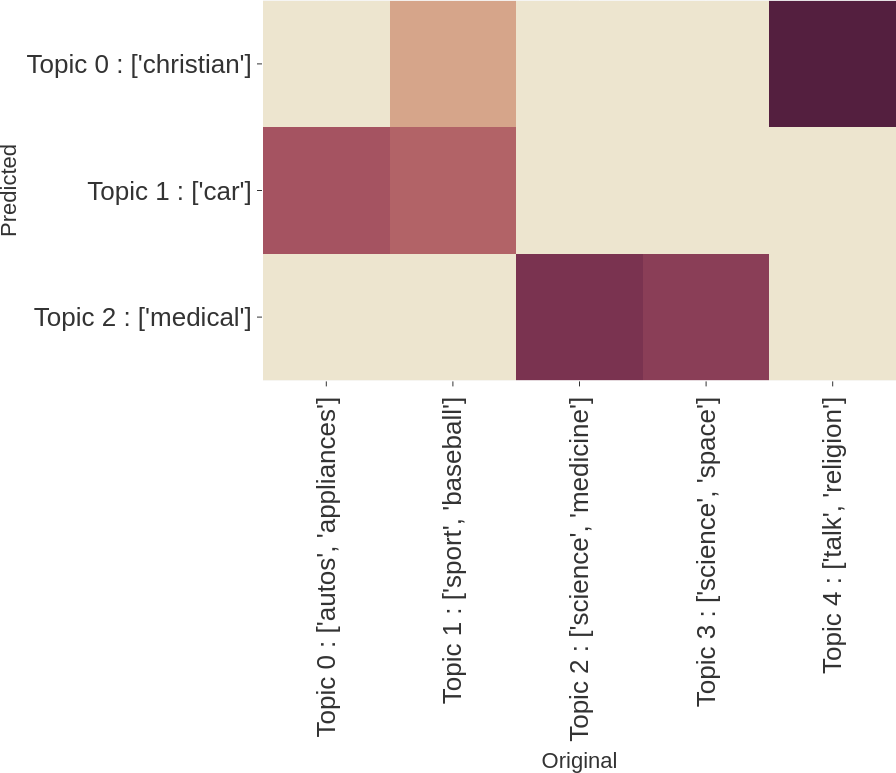}
        \label{fig:ot}
    }
    % \hspace{0.5cm}
    \subfigure[Cosine Distance]{
        \includegraphics[width=0.45\columnwidth]{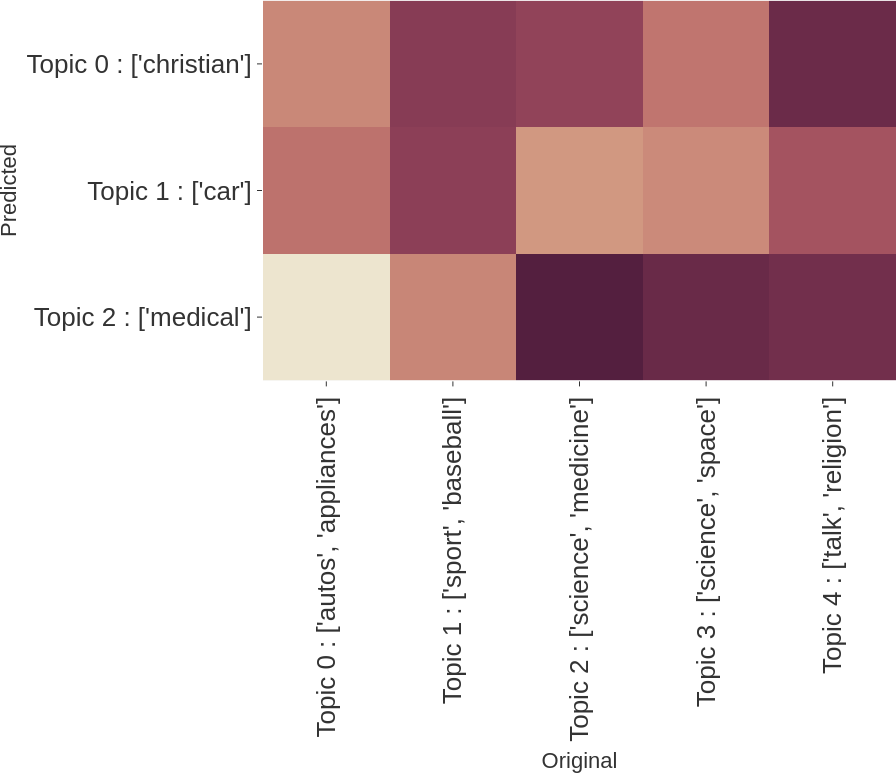}
        \label{fig:cosine}
    }
    \caption{The left figure shows the transport map, while the right one depicts the cosine similarity map. In both cases, darker cells indicate regions of higher transported mass, \cref{fig:ot}, or shorted cosine distance, \cref{fig:cosine}.}
    \label{fig:evaluation_ot}
\end{figure}
\begin{table}[htb!]
\centering
\caption{Topic merging and discovery accuracies on 50 simulated datasets. MA stands for Merging Accuracy, and DA for Discovery Accuracy.}
\begin{tabular}{l|c|c|c} 
\toprule
Method & MA & DA & $H_{(\text{MA}, \text{DA})}$\\ 
\midrule
UOT Cosine & 0.79 $\pm$ 0.24 & \textbf{0.93} $\pm$ 0.18 & \textbf{0.85}\\ 
UOT Euclidean & 0.75 $\pm$ 0.23 & 0.85 $\pm$ 0.28 & 0.79\\ 
UOT Minkowsky & 0.75 $\pm$ 0.23 & 0.85 $\pm$ 0.28 & 0.79\\ 
\midrule
CD & \textbf{0.84} $\pm$ 0.21 & 0.72 $\pm$ 0.26 & 0.77\\ 
ED & 0.74 $\pm$ 0.24 & 0.84 $\pm$ 0.25 & 0.76\\ 
\bottomrule
\end{tabular}
\label{tab:acc_disc}
\end{table}

Finally, we evaluate UOT and CD to align the predicted topics with the true topics provided by the dataset labels of \texttt{20kNewsGroup}. Specifically, we focus on the \textsc{Custom} setting shown in~\cref{fig:custom}.
Since ETMs treat documents as bag-of-words, we construct topic embeddings for what we refer to as `pure documents'—documents whose topics are directly derived from the dataset labels. Typically, these labels contain only two or three words, so we augment them by adding the most semantically similar words based on the GloVe model. Therefore, given a topic $k$, we can compute $\beta_k$ by considering the words in our `pure documents', and we can approximate the corresponding topic embeddings as 
\begin{align}
    \alpha_k \approx (\bm{\rho}^{\top})^{+} \ln(\beta_k),
\end{align}
where $(\bm{\rho}^{\top})^{+}$ is the Moore-Penrose pseudoinverse of $\bm{\rho}^{\top}$. The logarithm accounts for the inversion of the softmax transformation. Let us denote as $\widehat{\bm{\alpha}}$ the latent topic matrix extracted from the model at time $T$ and $\bm{\alpha}_{pure}$ the matrix computed from the `pure documents'. Our goal is to visualize how optimal transport and cosine similarity align $\widehat{\bm{\alpha}}$ with $\bm{\alpha}_{pure}$. To compute the transport map, we follow the same strategy used during training (\cref{sec:stream_etm}).

The results are shown in~\cref{fig:evaluation_ot}, where darker cells in the matrices indicate regions of higher transported mass (\cref{fig:ot}) or shorter Cosine Distance (\cref{fig:cosine}). As observed, the associations in the transport map appear reasonable, as the matrix is not particularly dense, and transportation occurs primarily between similar topics. An interesting case is the predicted topic 2, which pertains to medicine and is mapped between the two original topics containing the word \texttt{science}. Conversely, the Cosine Distance matrix appears denser, leading to less immediate associations.
For example, \texttt{christianity} could be associated with any original topic, even though the distance is slightly smaller from the topic \texttt{talk}, \texttt{religion}. This can lead to greater instability, as the minimum distance may not necessarily correspond to the correct topic from a human perspective.

\subsection{Qualitative analysis of the recovered topic dynamics}
\begin{figure}[tbh!]
    \centering
    \subfigure[Original]{\includegraphics[width=0.45\columnwidth]{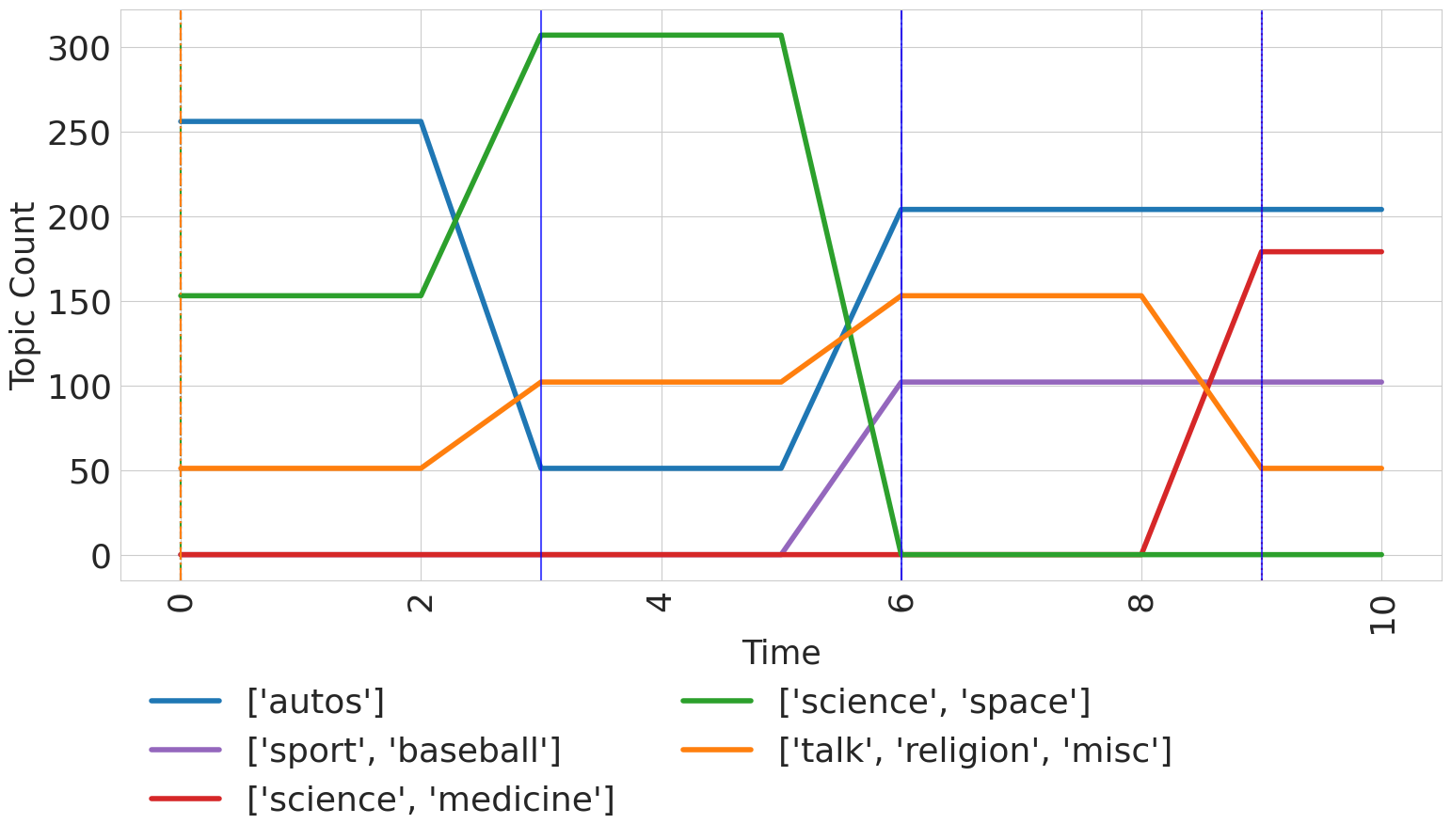}
        \label{fig:original_cs}
    }
    \hfill
    \subfigure[StreamETM]{
        \includegraphics[width=0.45\columnwidth]{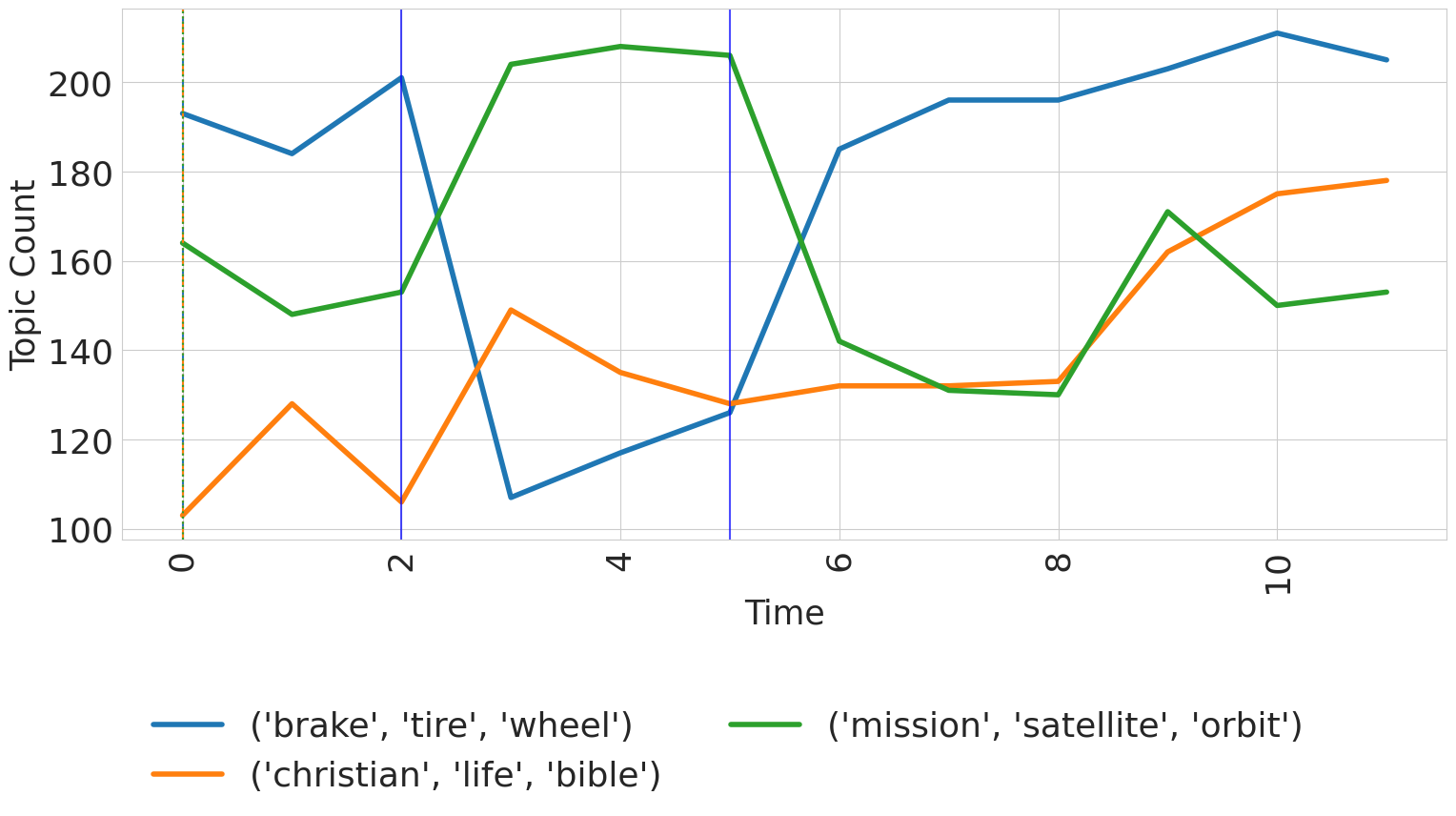}
        \label{fig:stream_cs}
    }
    \hfill
    \subfigure[MergeBERT]{
        \includegraphics[width=.9\columnwidth]{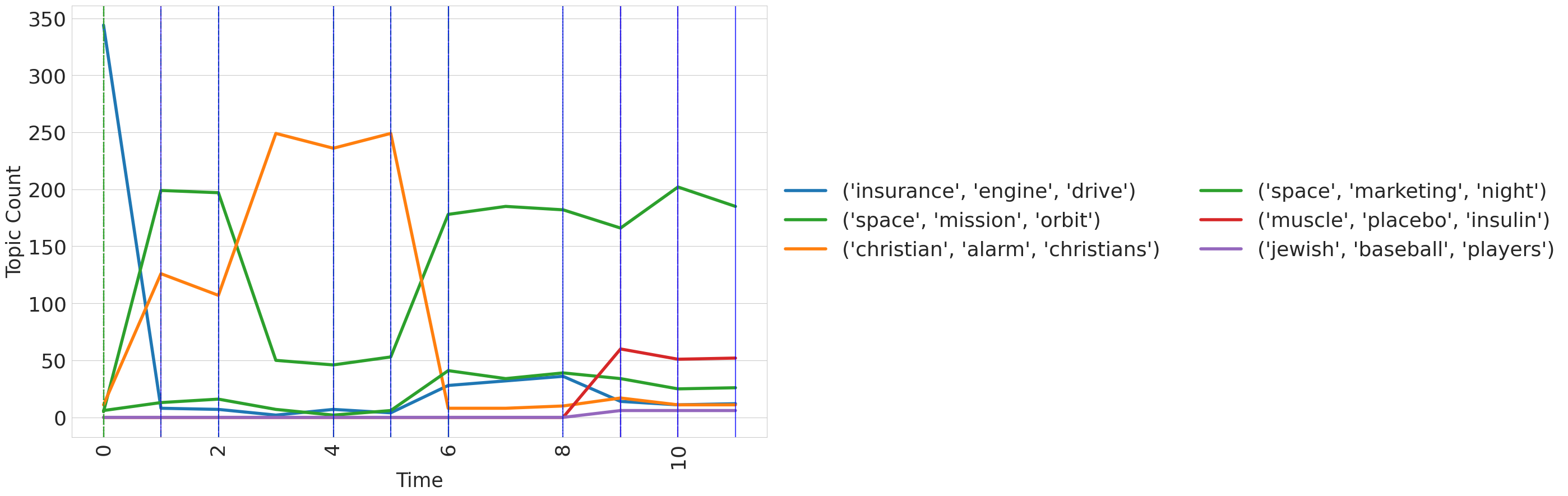}
        \label{fig:merge_cs}
    }
    \caption{Qualitative assessment of topic evolution over time in the \textsc{Custom} setting. Blue vertical lines indicate the change points detected by the algorithm.}
    \label{fig:custom}
\end{figure}
\begin{figure}[htb!]
    \centering
    \subfigure[StreamETM]{\includegraphics[width=0.45\columnwidth]{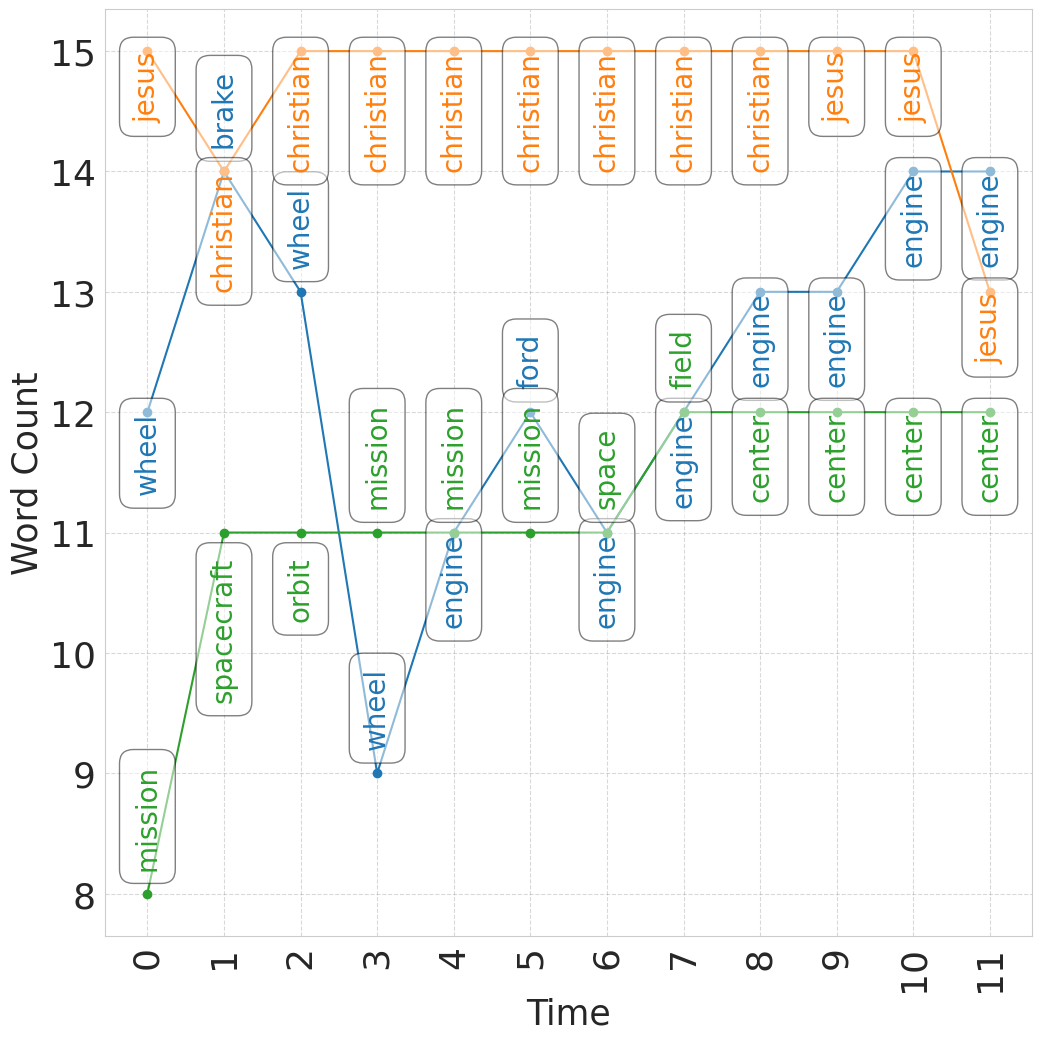
    }
    \label{fig:stream_wc_cs}
    }
    \hfill
    \subfigure[MergeBERT]{
    \includegraphics[width=0.45\columnwidth]{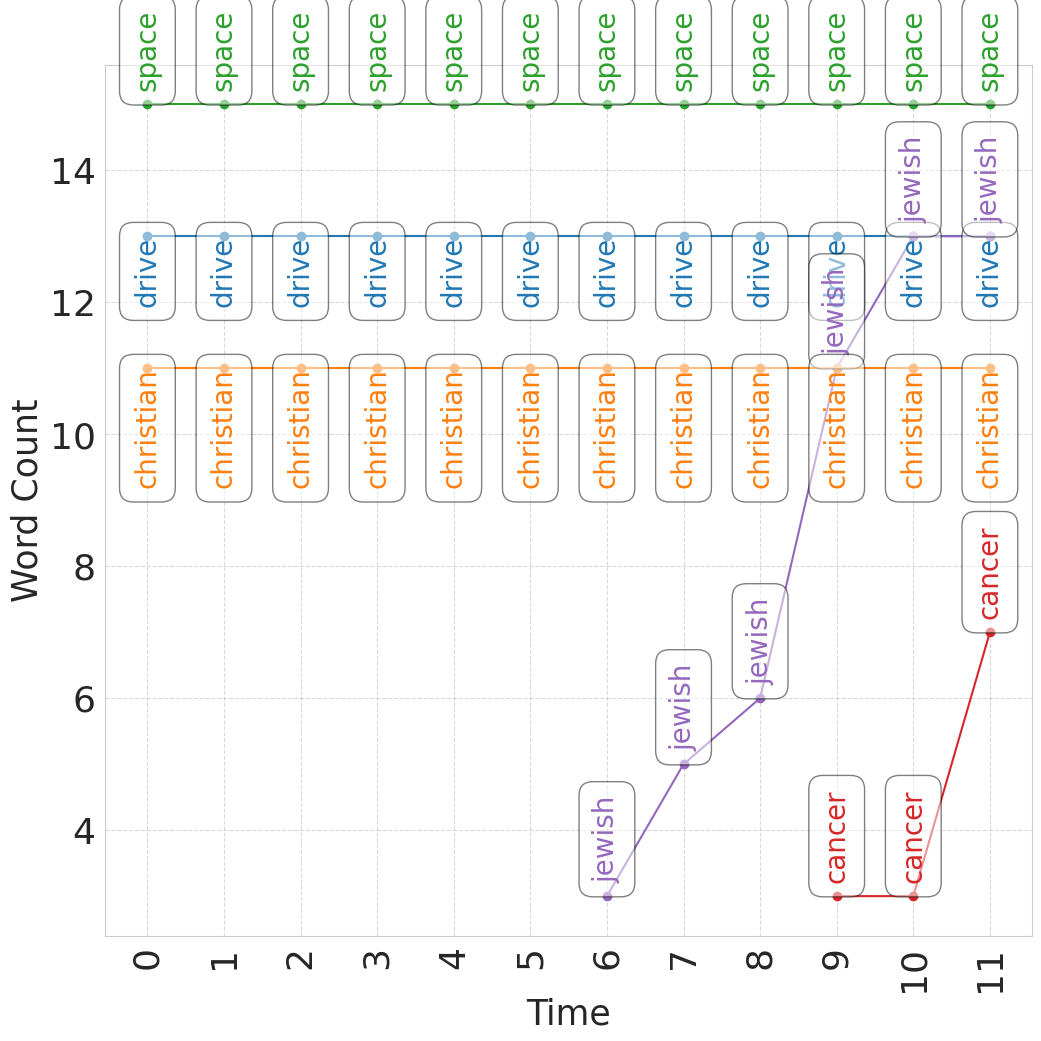}
    \label{fig:merge_wc_cs}
    }
    \caption{\textsc{Custom} setting. The most frequent word for topic across the 15 training runs.}
    \label{fig:custom_results}
\end{figure}

\begin{figure}
    \centering
    \subfigure[H$_{(TC, TD)}$]
    {\includegraphics[width=.45\columnwidth]{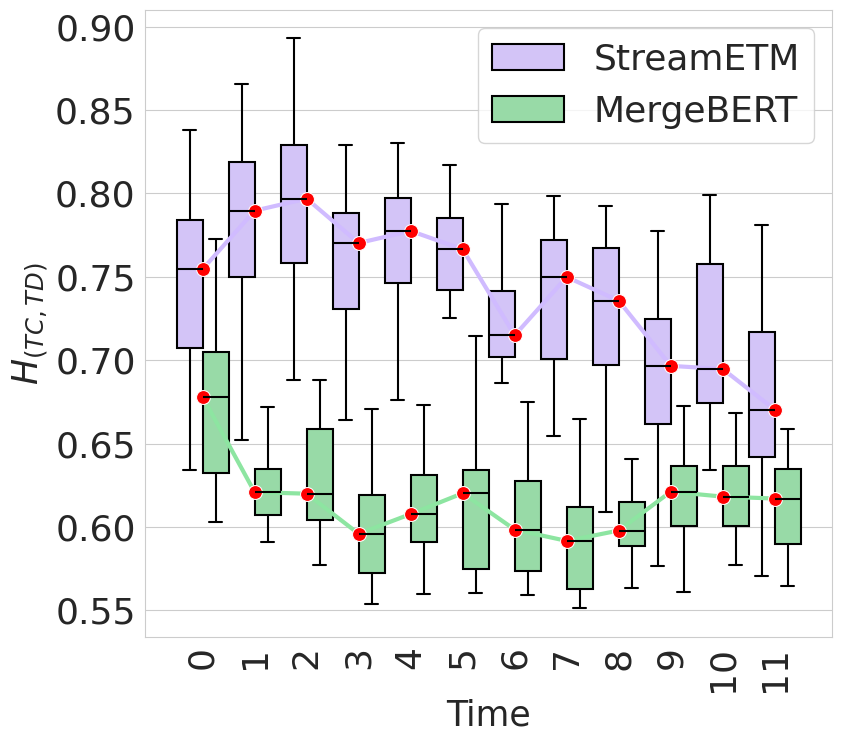}
     \label{fig:harmonic_cs}
    }
    \hfill
    \subfigure[OCPD]
    {\includegraphics[width=.45\columnwidth]{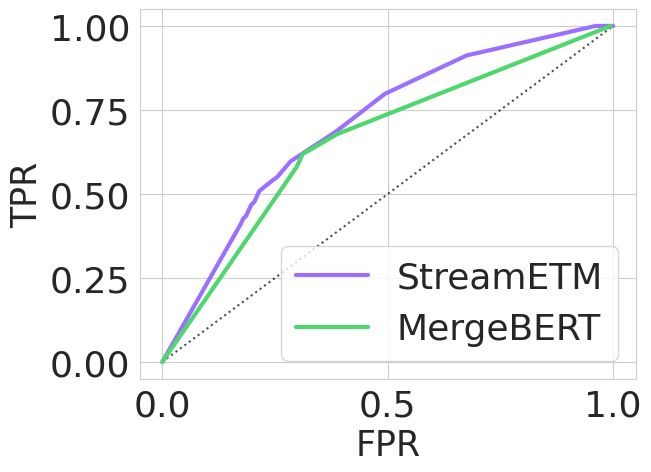}
     \label{fig:cp_cs}
    }
    \caption{\textsc{Custom} setting. In (a) harmonic mean between TC and TD and in (b) ROC curves. Results were computed across the 15 training runs.}
    \label{fig:custom_cpd_harmonic}
\end{figure}
\begin{figure}[htb!]
    \centering
    \subfigure[Original]{\includegraphics[width=0.45\columnwidth]{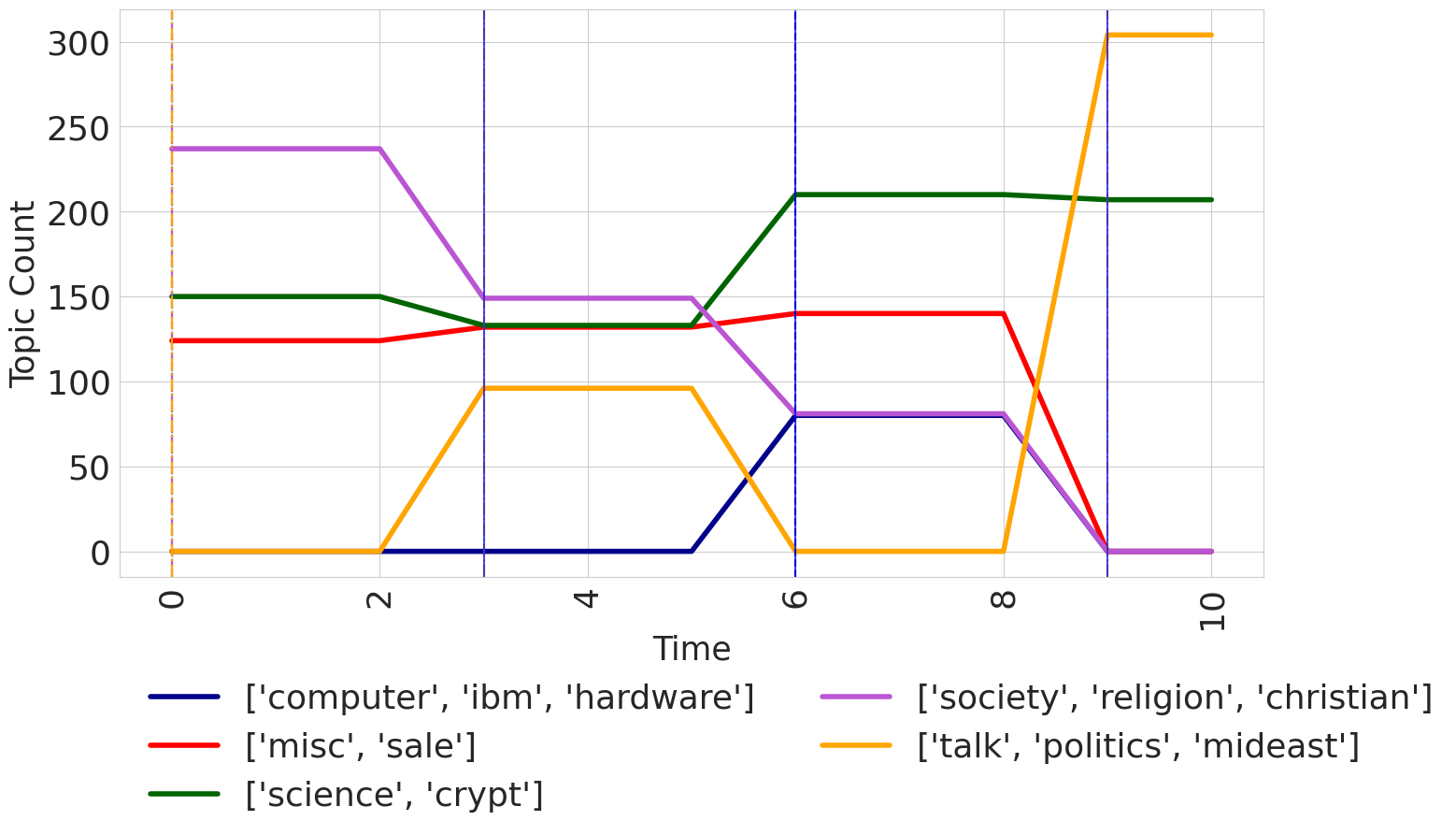}
        \label{fig:original_ex}
    }
    \hfill
    \subfigure[StreamETM]{
        \includegraphics[width=0.45\columnwidth]{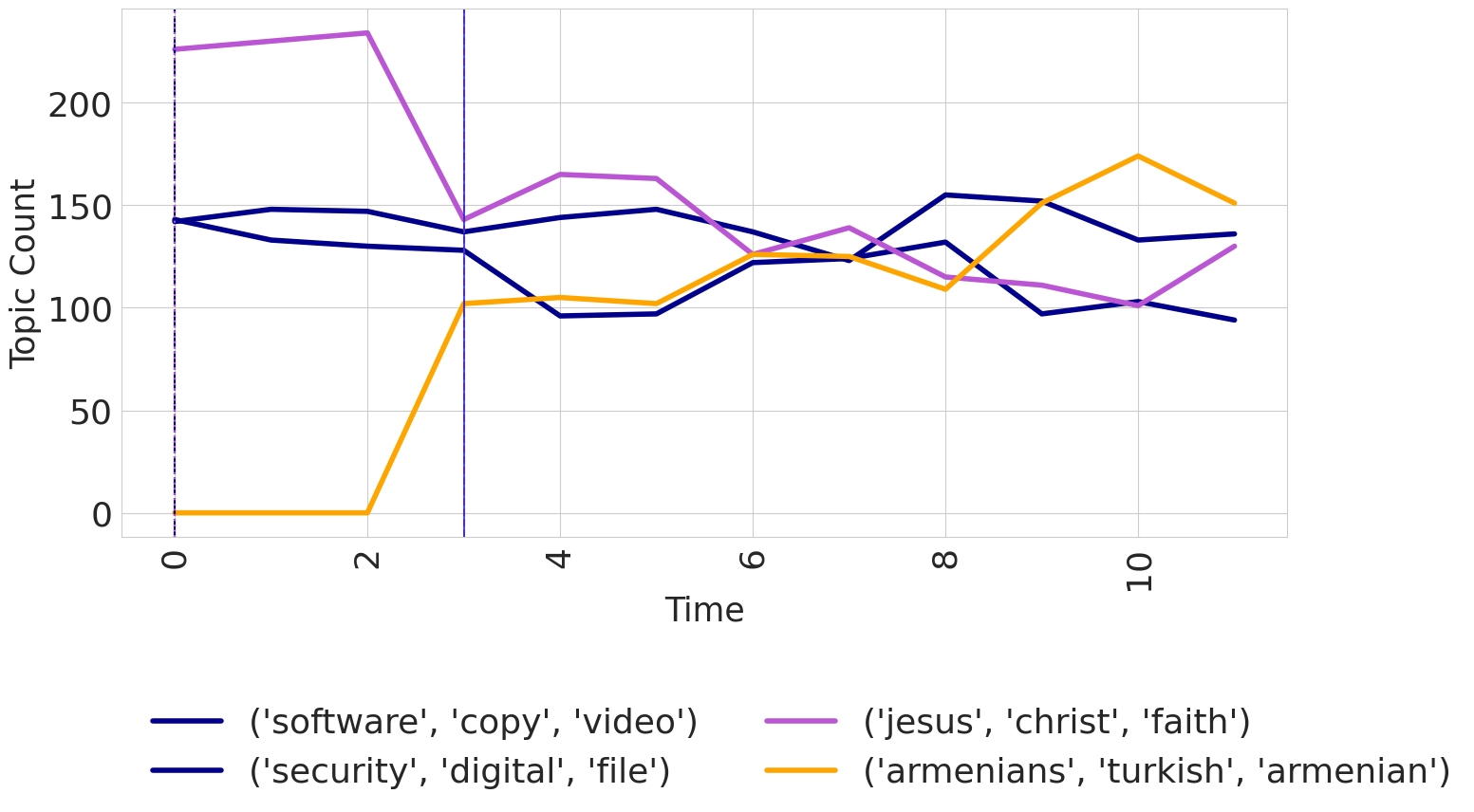}
        \label{fig:stream_ex}
    }
    \hfill
    \subfigure[MergeBert]{
        \includegraphics[width=.9\columnwidth]{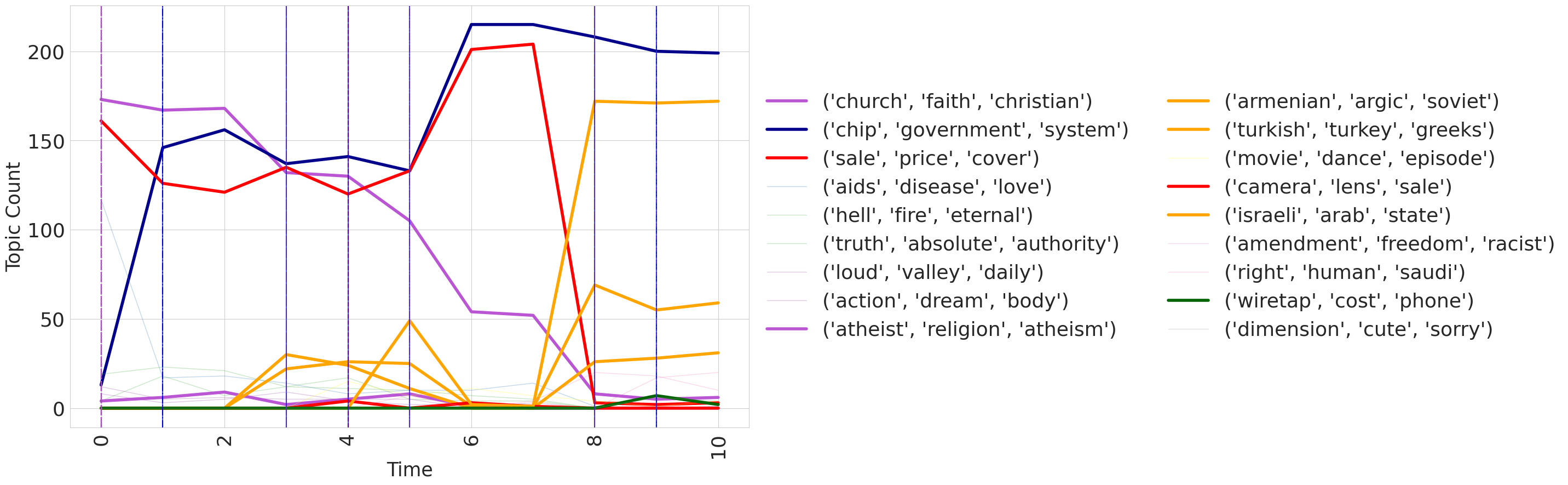}
        \label{fig:merge_ex}
    }
    \caption{Qualitative assessment ot topic evolution over time in the \textsc{Dynamic} setting. Blue vertical lines indicate the change points detected by the algorithm.}
    \label{fig:extreme}
\end{figure}
\begin{figure}[htb!]
    \centering
    \subfigure[StreamETM]{\includegraphics[width=0.45\columnwidth]{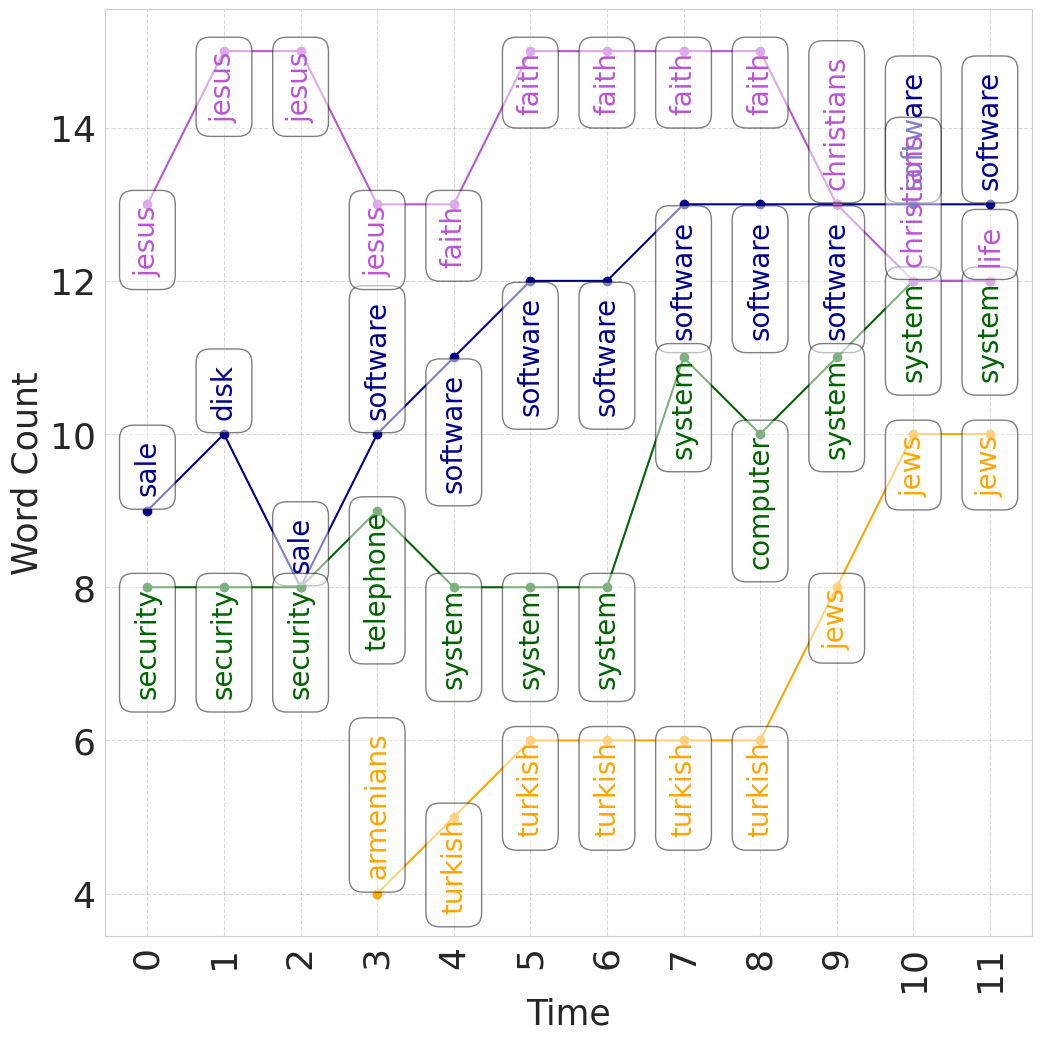
    }
    \label{fig:stream_wc_ex}
    }
    \hfill
    \subfigure[MergeBERT]{
    \includegraphics[width=0.45\columnwidth]{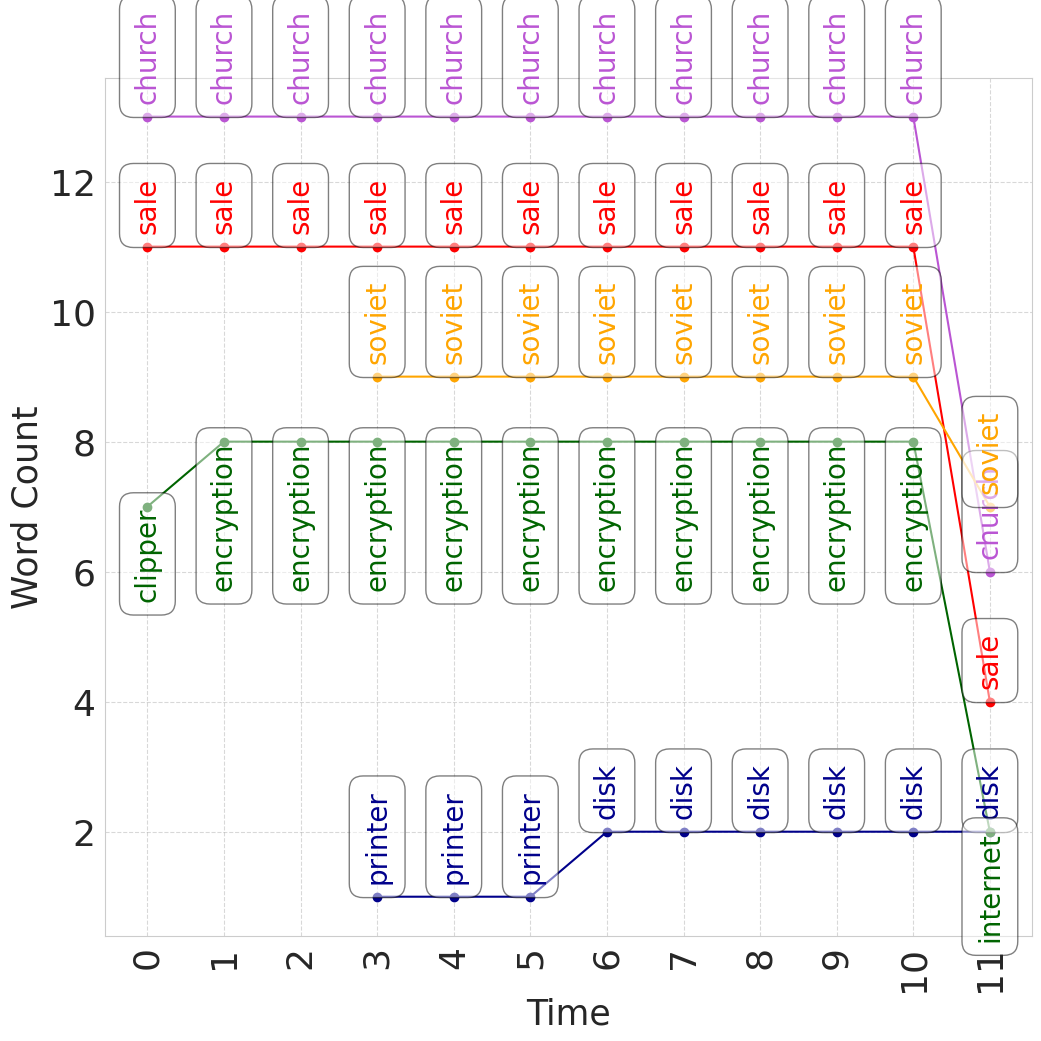}
        \label{fig:merge_wc_ex}
    }
    % \hfill
    % \subfigure[H$_{(TC, TD)}$]{
    % \includegraphics[width=0.3\columnwidth]{figs/imgs/extreme_harmonic.png}
    % \label{fig:harmonic_ex}
    % }
    \caption{\textsc{Dynamic} setting. The most frequent word for topic across the 15 training runs.}
    \label{fig:extreme_results}
\end{figure}

\begin{figure}
    \centering
    \hfill
    \subfigure[H$_{(TC, TD)}$]{
    \includegraphics[width=0.45\columnwidth]{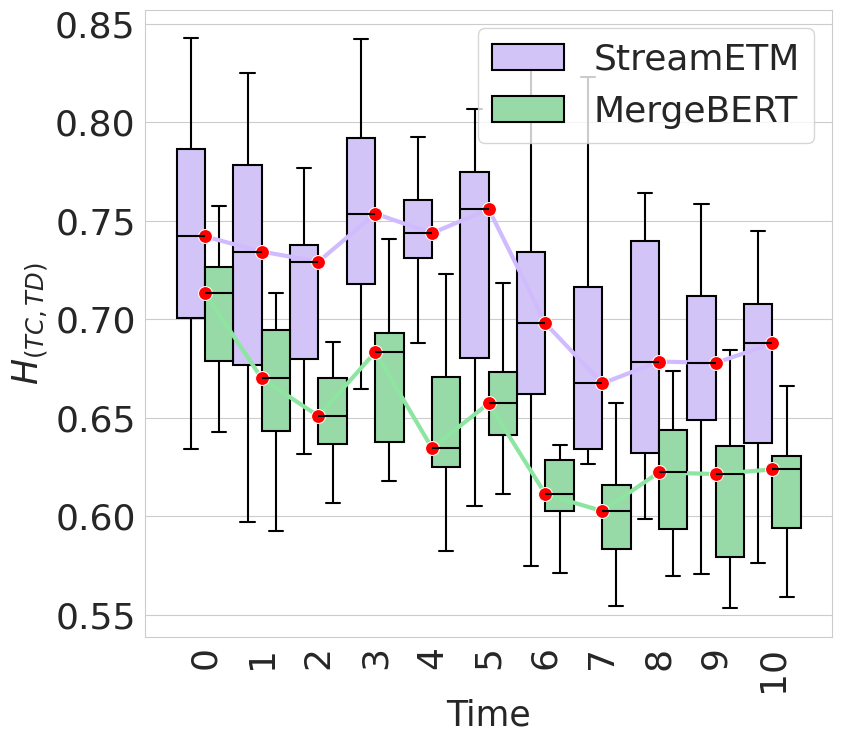}
    \label{fig:harmonic_ex}
    }
    \hfill
    \subfigure[OCPD]
    {\includegraphics[width=.45\columnwidth]{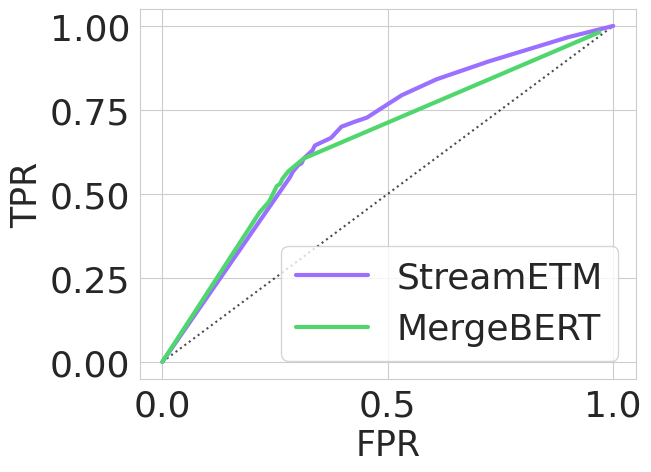}
     \label{fig:cp_ex}
    }
    \caption{\textsc{Dynamic} setting. In (a) harmonic mean between TC and TD and in (b) ROC curves. Results were computed across the 15 training runs.}
    \label{fig:extreme_cpd_harmonic}
\end{figure}
In~\cref{fig:custom,fig:extreme}, we plot the topic evolution over time for a randomly selected training run in the \textsc{Custom} and \textsc{Dynamic} settings, respectively. Even if the proposed StreamETM model cannot identify all five topics, it can mimic the original topics' evolutions. In~\cref{fig:stream_cs}, after time 6, the model likely merges the topics \texttt{science}, \texttt{medicine}, \texttt{sport}, and \texttt{baseball} with other topics. Specifically, the \texttt{space} topic, instead of disappearing, likely absorbs the \texttt{medicine} topic, as both could contain similar terms. In~\cref{fig:merge_cs}, we observe a similar shape as in~\cref{fig:original_cs}, but the topics are swapped (cf. \texttt{space} and \texttt{religion}, in MergeBERT,  vs. \texttt{autos}, \texttt{space}, in the original distribution).
As can be observed from the plots, compared to StreamETM, MergeBERT generates an excessive number of topics, leading to an overproliferation of topics. For example, in~\cref{fig:merge_cs} the topic \texttt{insurance}, \texttt{engine}, \texttt{drive} could be regarded as the same topic as \texttt{radar}, \texttt{tire}, \texttt{detector}. More evidently, in~\cref{fig:merge_ex}, \texttt{church}, \texttt{faith}, \texttt{christian} and \texttt{atheist}, \texttt{religion}, \texttt{atheism}. MergeBERT appears to capture more subtopics, while StreamETM focuses on larger concepts. Several factors contribute to this difference: (i) MergeBERT is based on a SentenceTransformer model, which captures more granular, context-dependent information, whereas StreamETM treats documents as a bag of words; (ii) MergeBERT’s performance is highly dependent on a large number of parameters, making it challenging and impractical to tune for each document; (iii) MergeBERT does not explicitly merge topics, instead selecting top words from the first model if cosine similarity is sufficiently high; (iv) in StreamETM, the UOT mechanism enables topic embeddings to merge more effectively, an adjustable threshold on the transported mass would allow for fewer merges and the creation of more distinct topics.

In~\cref{fig:stream_wc_cs,fig:merge_wc_cs}, in~\cref{fig:stream_wc_ex,fig:merge_wc_ex}, we display the most frequent word for each topic across the 15 training runs and manually align the topic indices across the different executions. Since MergeBERT identifies more than 20 topics, we focus on the 5 topics of particular interest to us. In the \textsc{Custom} setting, we observe that for both models, the most frequent word is \texttt{christian}, and both models show high stability around this topic, with that word appearing in almost all 15 training runs. However, we see much more variability in the other two topics in~\cref{fig:stream_wc_cs}, reflecting the evolution of topics over time. For instance, the frequency of the word \texttt{wheel} associated with the \texttt{autos} topic diminishes at time 3. A similar trend is observed in the \textsc{Dynamic} setting. As mentioned earlier, one of the strengths of StreamETM is its ability to adapt topics over time as the text corpus evolves. The 5 top words for each topic over time across the training executions are provided in Tabs. 4 and 5 (cf. Appendix A.3). %~\cref{tab:topics_custom,tab:topics_extreme}.

\subsection{Quantitative analysis of the recovered topic dynamics}
\Cref{fig:harmonic_cs,fig:harmonic_ex} illustrate the harmonic mean between TC and TD, denoted as $H_{(TC, TD)}$, across the 15 training executions. The numerical results are provided in Tabs. 2 and 3 (cf. Appendix A.2). %~\cref{tab:harmonic_custom,tab:harmonic_extreme}. 
From the TD perspective, both models achieve satisfactory results. However, more significant differences are observed from the TC perspective. Notably, when comparing the box plot in~\cref{fig:harmonic_cs} with the topic evolution in~\cref{fig:custom}, we observe a decrease in the metric at time 6, which corresponds to the moment when there is an inversion in the distribution between the blue and green topics, alongside the emergence of the new violet topic. Following this, the metric stabilizes until time 9, when the new red topic is introduced. A similar pattern is evident in~\cref{fig:harmonic_ex}. This behavior is expected, as the model does not identify the new topic at the time step, leading to decreased topic coherence. Finally, MergeBERT shows the same behavior as StreamETM in~\cref{fig:harmonic_ex}, but at a lower TC, instead in~\cref{fig:harmonic_cs}, the behavior is slightly inverted, and this could be given to the fact that from time 0 to 6 the model associates the original blue topic evolution to the green one and the green with the orange one.

\subsection{Online change point detection}
\Cref{fig:cp_cs,fig:cp_ex} present the performance of the OCPD algorithm in terms of ROC curves across the 15 training runs. If the topic of evolution is correctly predicted, the algorithm should detect rupture points at approximately the same time steps. The results show that this final task is extremely difficult, and both methods exhibit certain drawbacks. With StreamETM, fewer topics are detected, which increases the likelihood that the OCPD algorithm will identify fewer rupture points than expected. In contrast, MergeBERT experiences an explosion in the number of topics, likely leading to more false positives.

\section{Conclusions}
\label{sec:conclusion}
We considered the extremely challenging problem of online topic modeling on document streams, with online change point detection. In order to address the limitations of existing online topic modeling approaches, we introduced StreamETM, an online extension of the Embedded Topic Model (ETM) for streams of text documents. Our method leverages variational inference to update topic distributions sequentially while incorporating optimal transport to monitor, merge, and discover evolving topics over time. This approach ensures that topics remain coherent despite the continuous influx of new documents.
Beyond model development, we complemented StreamETM with a change point detection algorithm to automatically identify shifts in topic dynamics. This enables the proposed approach to provide a synthetic summary of the document stream through intelligible topics and to issue alerts to the analysts when significant changes in the dynamics of text streams are detected. Our experiments demonstrate that StreamETM effectively adapts to streaming textual data. Optimal transport provides a principled way to associate topics across time windows, addressing the shortcomings of static clustering methods. Extensive numerical experiments in simulated and real-world scenarios showed that StreamETM outperforms its competitors in various scenarios. Future work includes modeling the extension of the vocabulary that may evolve over time, and proposing a model selection strategy to determine on the fly the appropriate number of topics.

\section{Acknowledgment}
This work has been supported by the French government through the 3IA Côte d’Azur Investments, which are managed by the National Research Agency (ANR) with the reference number ANR-23-IACL-0001.

\appendix
\section{Supplementary Materials}

\subsection{OnlineBERT}
\label{app:online}
\begin{figure}[htb!]
    \centering
    \subfigure[Top 5 topics]{\includegraphics[width=0.45\columnwidth]{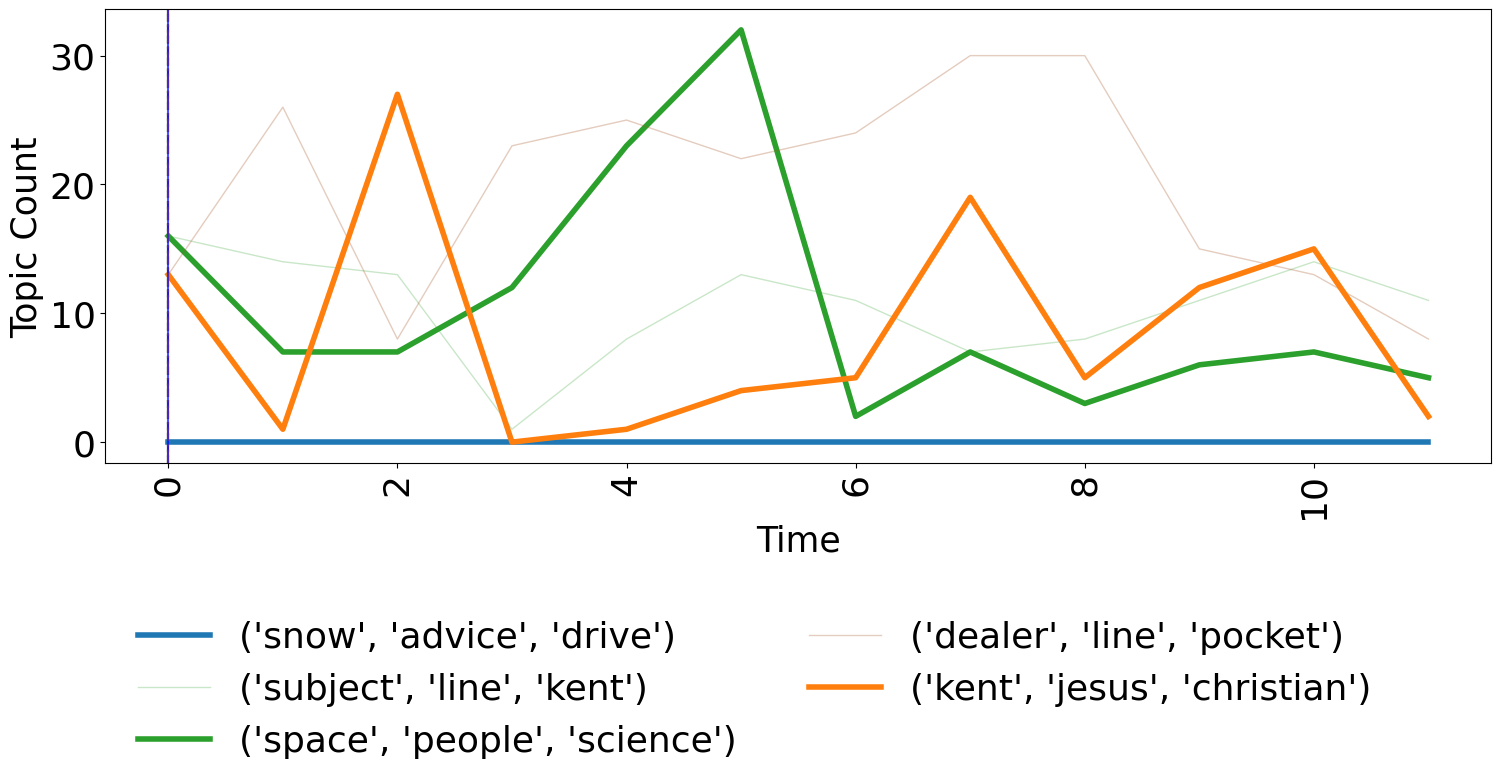}
        \label{fig:online_5_cs}
    }
    \hfill
    \subfigure[Top 10 topics]{
        \includegraphics[width=0.45\columnwidth]{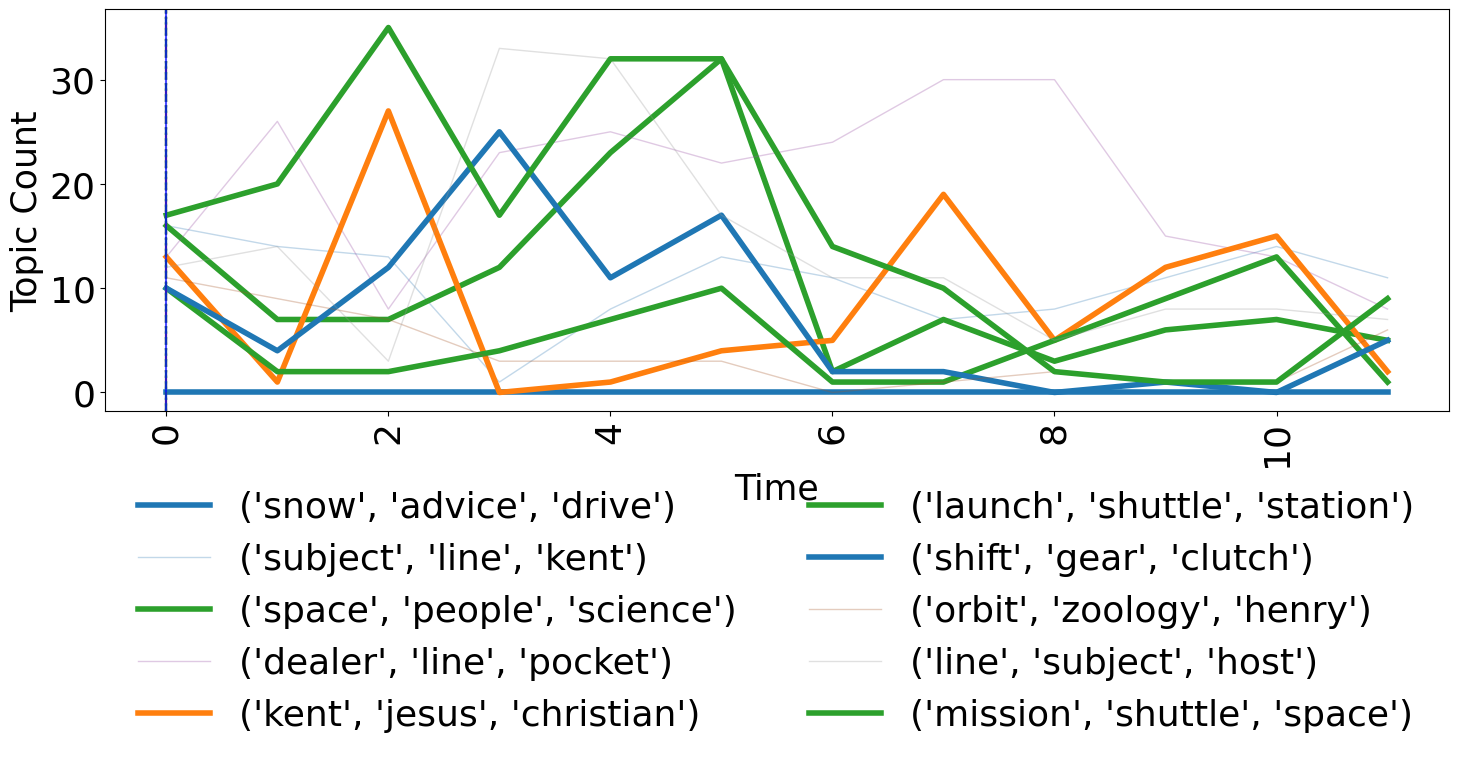}
        \label{fig:online_10_cs}
    }
        \hfill
    \subfigure[Top 30 topics]{
    \includegraphics[width=.95\columnwidth]{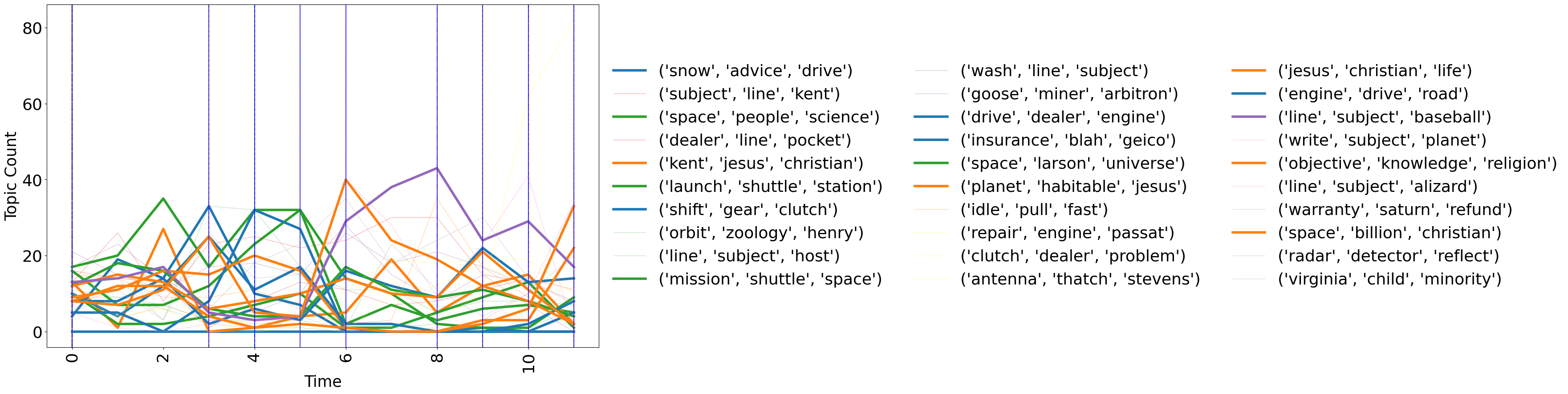}
        \label{fig:online_30_cs}
    }
    \hfill
    \subfigure[All topics]{
        \includegraphics[width=.95\columnwidth]{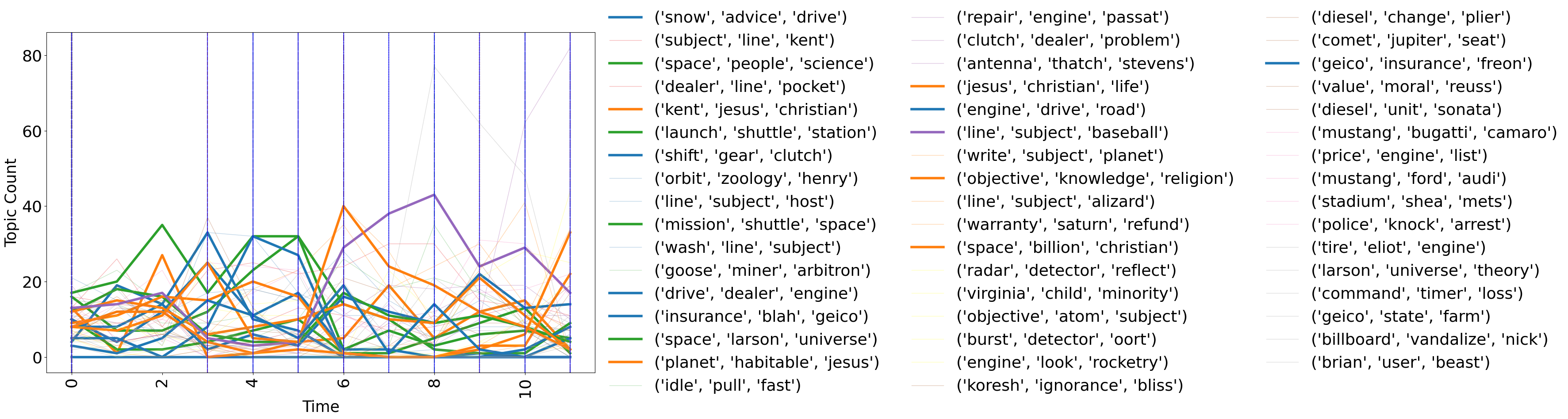}
        \label{fig:online_50_cs}
    }
    \caption{Qualitative assessment \textsc{Custom} setting, OnlineBERT. Topic evolution over time. Blue markers indicate the change points detected by the algorithm.}
    \label{fig:custom_online}
\end{figure}

\begin{figure}[htb!]
    \centering
    \subfigure[Top 5 topics]{\includegraphics[width=0.45\columnwidth]{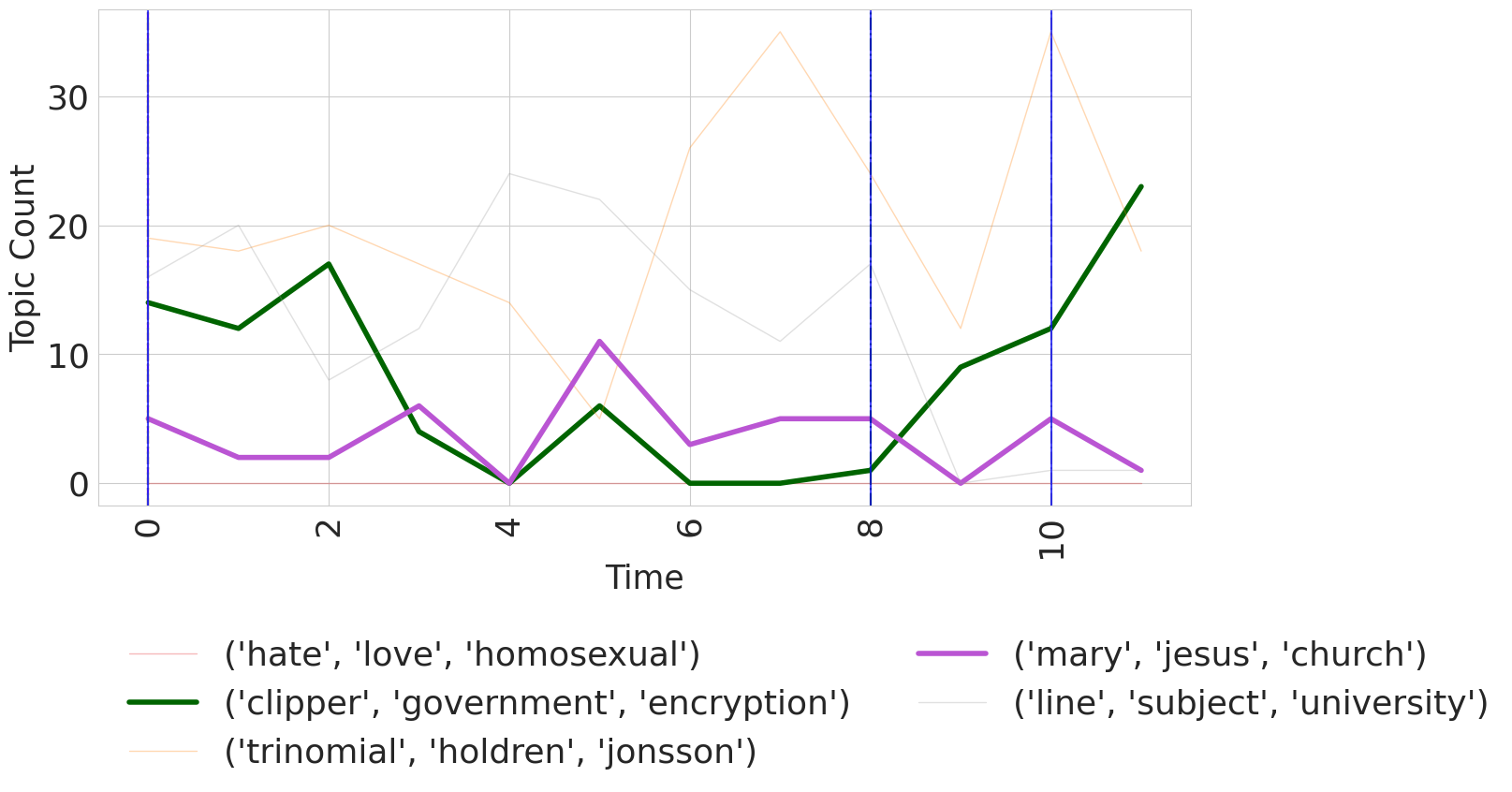}
        \label{fig:online_5_ex}
    }
    \hfill
    \subfigure[Top 10 topics]{
        \includegraphics[width=0.45\columnwidth]{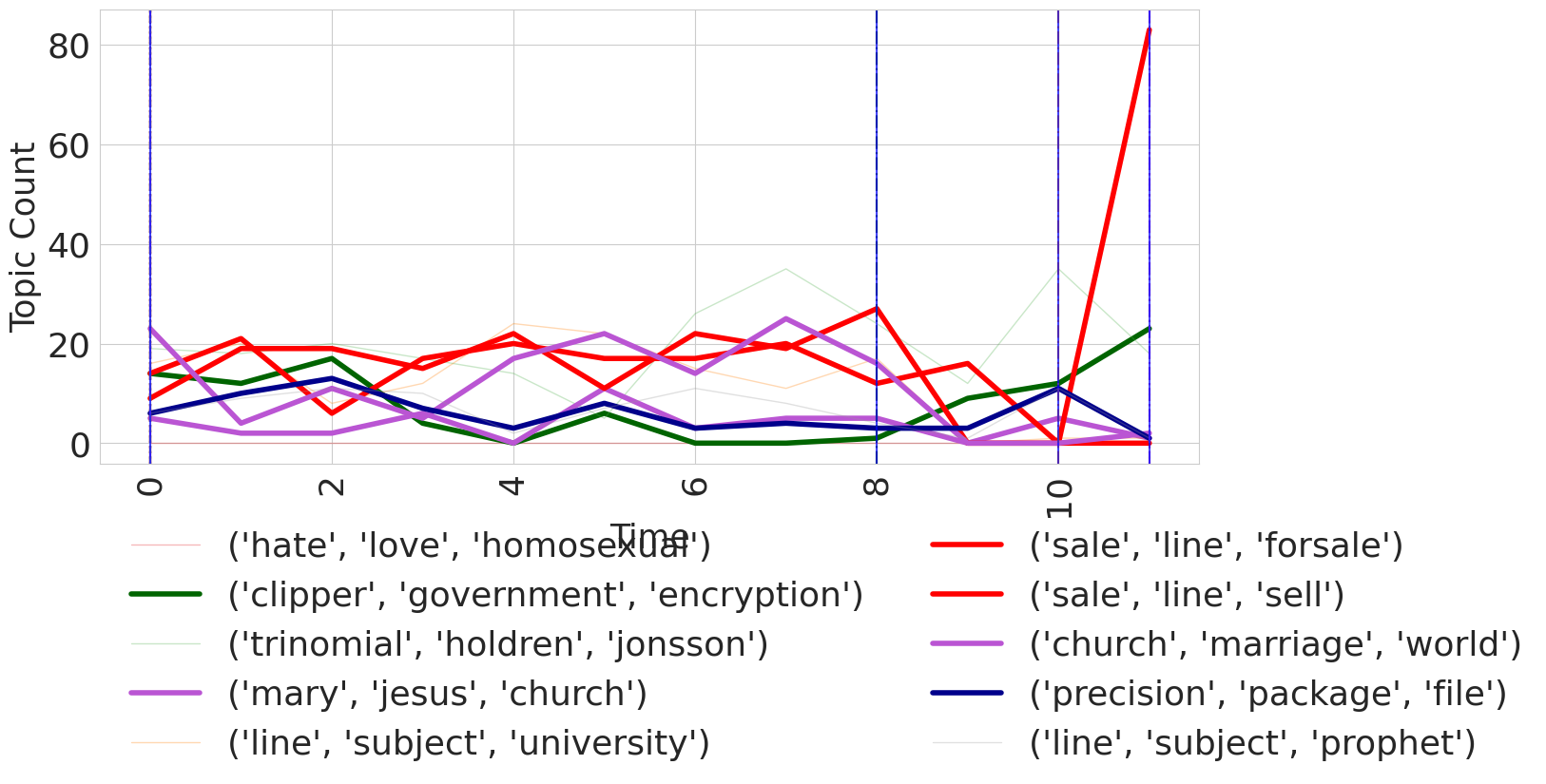}
        \label{fig:online_10_ex}
    }
        \hfill
    \subfigure[Top 30 topics]{
    \includegraphics[width=.95\columnwidth]{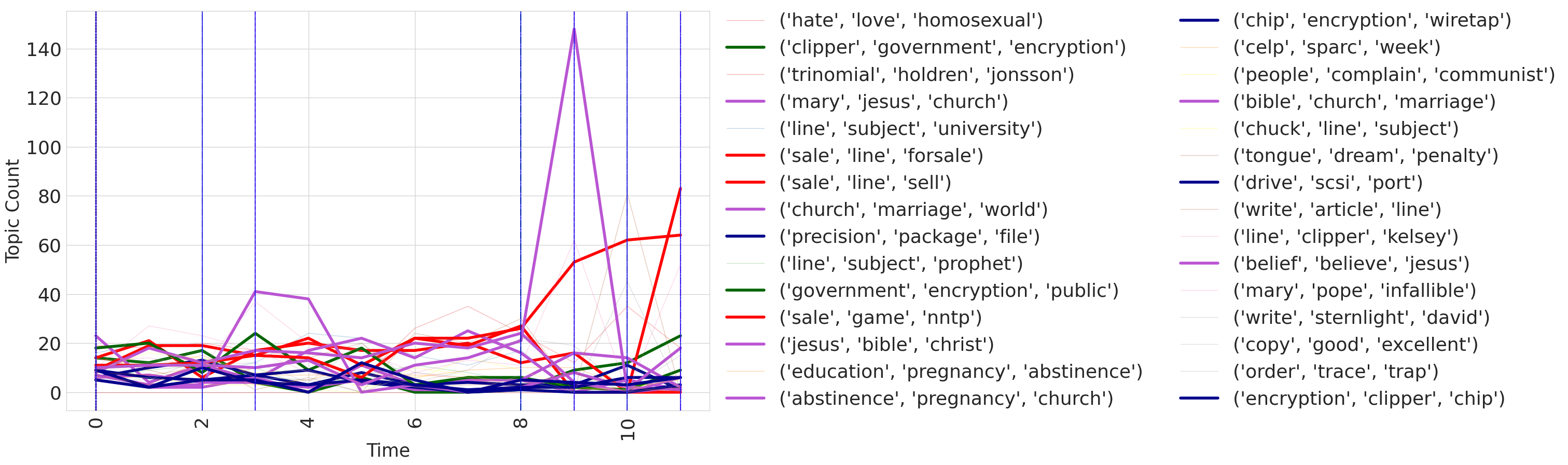}
        \label{fig:online_30_ex}
    }
    \hfill
    \subfigure[All topics]{
        \includegraphics[width=.95\columnwidth]{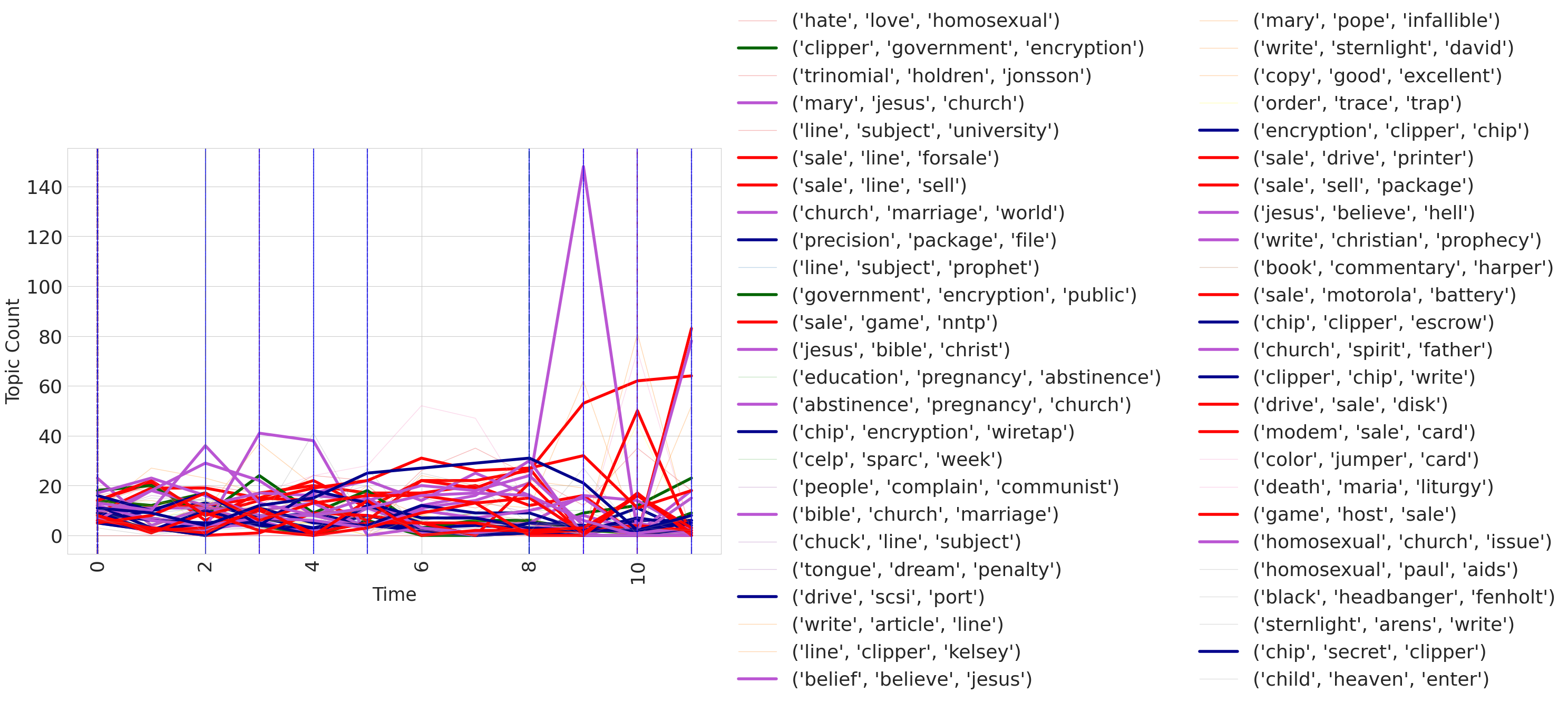}
        \label{fig:online_50_ex}
    }
    \caption{Qualitative assessment \textsc{Dynamic} setting, OnlineBERT. Topic evolution over time. Blue markers indicate the change points detected by the algorithm.}
    \label{fig:extreme_online}
\end{figure}

We used the OnlineBERT version as provided in \url{https://maartengr.github.io/BERTopic/getting_started/online/online.html}, where the model is tested against the same dataset. Specifically, OnlineBERT outputs the number of topics equal to the number of clusters (50), meaning that the number of topics cannot evolve. In~\cref{fig:custom_online,fig:extreme_online}, we present different executions of the model, considering the top 5, 10, and 30 topics. Providing the model with the \textit{exact} number of topics would introduce additional information not available to StreamETM and MergeBERT. However, to reduce unnecessary noise from the full set of 50 topics, we report results based on the 10 most important topics throughout this section.

\begin{figure}[htb!]
    \centering
    \subfigure[StreamETM]{\includegraphics[width=0.3\columnwidth]{figs/imgs/topics_over_executions_custom_stream.png
    }
    % \label{fig:stream_wc_ex}
    }
    \hfill
    \subfigure[OnlineBERT]{
    \includegraphics[width=0.3\columnwidth]{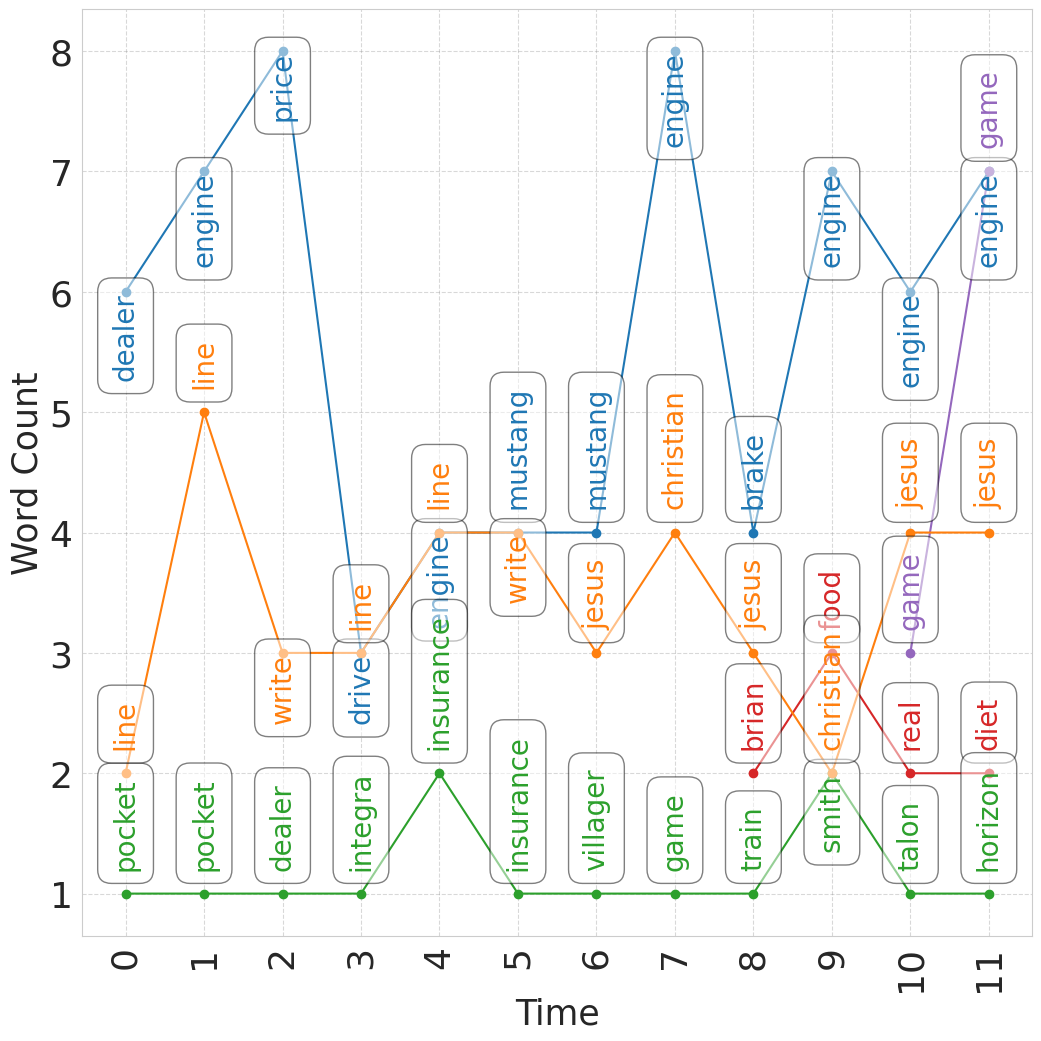}
        \label{fig:online_wc_cs}
    }
    \hfill
    \subfigure[H$_{(TC, TD)}$]{
    \includegraphics[width=0.3\columnwidth]{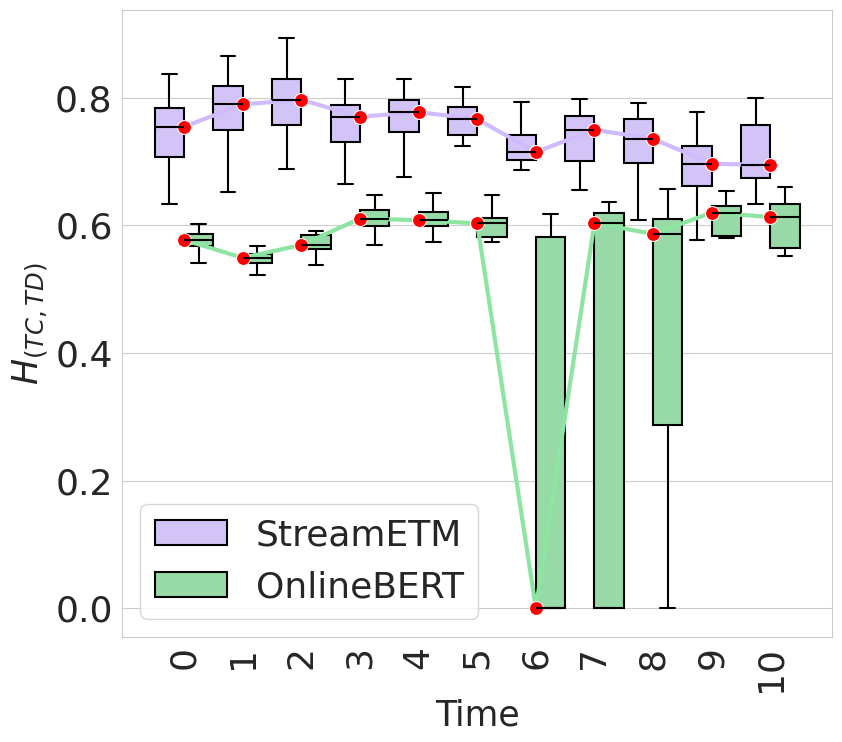}
    \label{fig:harmonic_cs_2}
    }
    \caption{\textsc{Custom} setting. In (a) and (b), the most frequent word for each topic; in (c) harmonic mean between TC and TD across the 15 training runs.}
    \label{fig:online_custom_reults}
\end{figure}

% \subsubsection{\textsc{Dynamic} setting.}

\begin{figure}[htb!]
    \centering
    \subfigure[StreamETM]{\includegraphics[width=0.3\columnwidth]{figs/imgs/topics_over_executions_extreme_stream.png
    }
    % \label{fig:stream_wc_ex}
    }
    \hfill
    \subfigure[OnlineBERT]{
    \includegraphics[width=0.3\columnwidth]{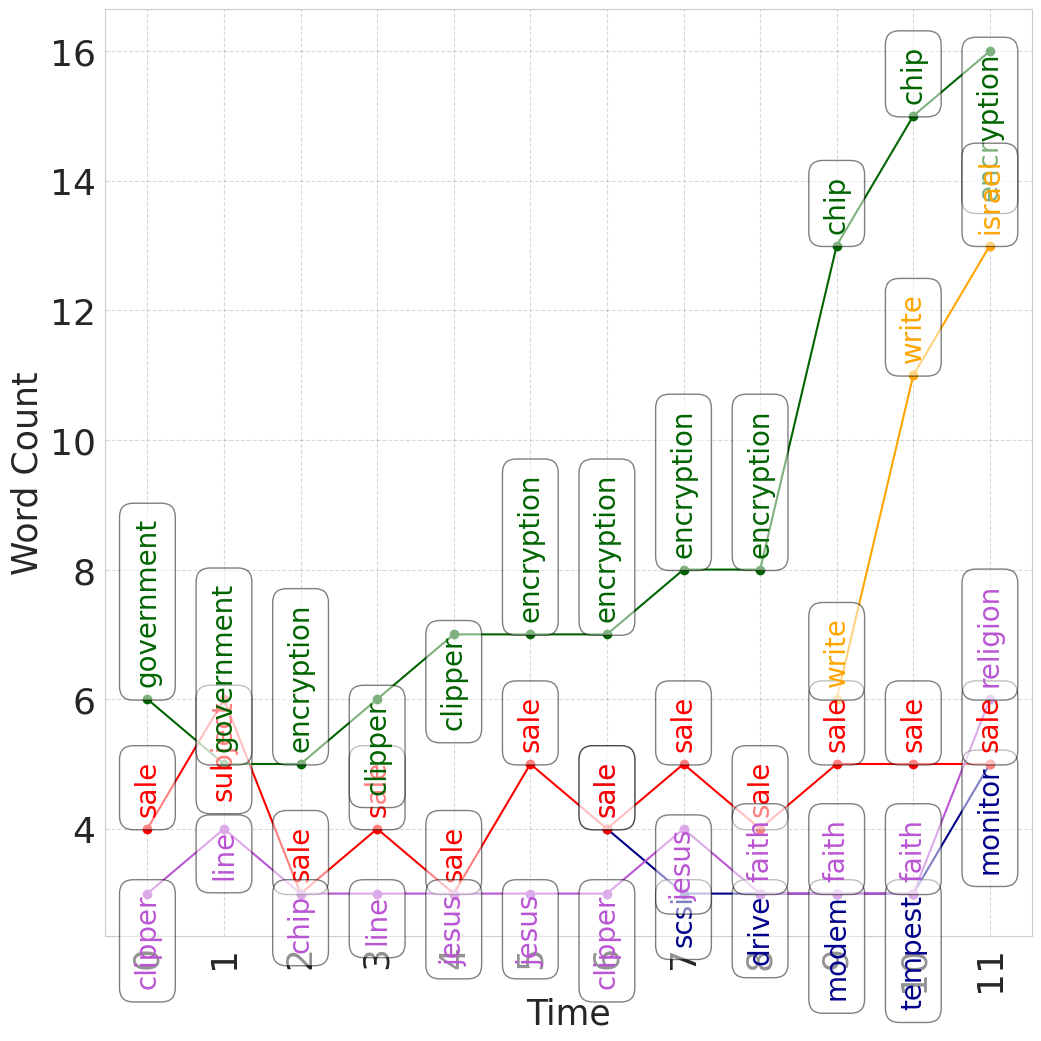}
        \label{fig:online_wc_ex}
    }
    \hfill
    \subfigure[H$_{(TC, TD)}$]{
    \includegraphics[width=0.3\columnwidth]{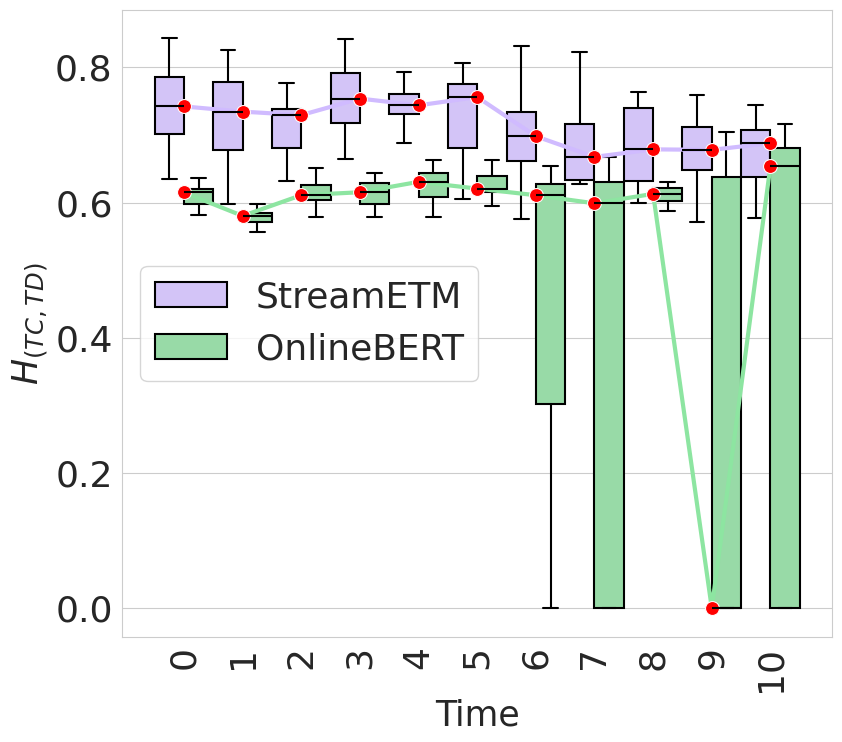}
    \label{fig:harmonic_ex_2}
    }
    \caption{\textsc{Dynamic} setting. In (a) and (b), the most frequent word for each topic; in (c) harmonic mean between TC and TD across the 15 training runs.}
    \label{fig:online_extreme_reults}
\end{figure}

\subsection{Additional numerical results}
\label{app:numerical}
This section provides the full numerical results for Topic Coherence and Topic Diversity (cf.~\cref{tab:harmonic_custom,tab:harmonic_extreme}).
\begin{table}[htb!]
    \centering
    \caption{\textsc{Custom} setting. TC stands for Topic Coherence, TD for Topic Diversity, and $H$ indicates the harmonic mean between Topic Coherence and Topic Diversity.}
    \resizebox{\columnwidth}{!}{
    \begin{tabular}{c|c|c|c||c|c|c||c|c|c}
    \toprule
    Time & \multicolumn{3}{c||}{\textsc{streamETM}} & \multicolumn{3}{c||}{\textsc{mergeBERT}} & \multicolumn{3}{c}{\textsc{onlineBERT}}\\
    \cmidrule{2-10}
         & TC & TD & H$_{(\text{TC,TD})}$ & TC & TD & H$_{(\text{TC,TD})}$
         & TC & TD & H$_{(\text{TC,TD})}$\\
    \midrule
         $t_{0}$ & 
         \textbf{0.59}$~\pm~$0.07 & \textbf{1.0}$~\pm~$0.01 & \textbf{0.74}$~\pm~$0.06 & 
         0.52$~\pm~$0.06 & 0.97$~\pm~$0.02 & 0.68$~\pm~$0.05 & 0.43$\pm$0.02 & 0.85$\pm$0.01 & 0.58$\pm$0.02\\
        $t_{1}$ & 
        \textbf{0.64}$~\pm~$0.08 & \textbf{1.0}$~\pm~$0.00 & \textbf{0.78}$~\pm~$0.06 & 
        0.46$~\pm~$0.02 & 0.97$~\pm~$0.02 & 0.62$~\pm~$0.02 & 
        0.44$\pm$0.02 & 0.73$\pm$0.02 & 0.55$\pm$0.02\\
        $t_{2}$ & 
        \textbf{0.65}$~\pm~$0.08 & \textbf{1.0}$~\pm~$0.00 & \textbf{0.79}$~\pm~$0.06 & 
        0.47$~\pm~$0.04 & 0.97$~\pm~$0.02 & 0.63$~\pm~$0.03 & 
        0.41$\pm$0.12 & 0.84$\pm$0.02 & 0.54$\pm$0.15\\
        $t_{3}$ & 
        \textbf{0.61}$~\pm~$0.06 & \textbf{1.0}$~\pm~$0.01 & \textbf{0.76}$~\pm~$0.05 & 
        0.43$~\pm~$0.04 & 0.97$~\pm~$0.02 & 0.60$ ~\pm~$0.03 & 
        0.45$\pm$0.13 & 0.85$\pm$0.02 & 0.57$\pm$0.16\\
        $t_{4}$ & 
        \textbf{0.62}$~\pm~$0.07 & \textbf{1.0}$~\pm~$0.01 & \textbf{0.76}$~\pm~$0.05 & 
        0.45$~\pm~$0.04 & 0.97$~\pm~$0.02 & 0.61$~\pm~$0.04 & 
        0.45$\pm$0.13 & 0.84$\pm$0.02 & 0.57$\pm$0.16\\
        $t_{5}$ & 
        \textbf{0.61}$~\pm~$0.05 & \textbf{1.0}$~\pm~$0.02 & \textbf{0.76}$~\pm~$0.04 & 
        0.45$~\pm~$0.05 & 0.97$~\pm~$0.02 & 0.61$~\pm~$0.04 & 
        0.41$\pm$0.17 & 0.85$\pm$0.02 & 0.53$\pm$0.21\\
        $t_{6}$ & 
        \textbf{0.56}$~\pm~$0.05 & \textbf{1.0}$~\pm~$0.01 & \textbf{0.72}$~\pm~$0.04 & 
        0.44$~\pm~$0.04 & 0.97$~\pm~$0.02 & 0.60$ ~\pm~$0.04 & 
        0.18$\pm$0.23 & 0.84$\pm$0.02 & 0.23$\pm$0.3 \\
        $t_{7}$ & 
        \textbf{0.59}$~\pm~$0.05 & \textbf{1.0}$~\pm~$0.01 & \textbf{0.74}$~\pm~$0.04 & 
        0.43$~\pm~$0.04 & 0.97$~\pm~$0.02 & 0.59$~\pm~$0.04 & 
        0.29$\pm$0.25 & 0.85$\pm$0.02 & 0.37$\pm$0.31\\
        $t_{8}$ & 
        \textbf{0.57}$~\pm~$0.06 & \textbf{1.0}$~\pm~$0.01 & \textbf{0.73}$~\pm~$0.05 & 
        0.43$~\pm~$0.03 & 0.97$~\pm~$0.02 & 0.60$ ~\pm~$0.03 & 
        0.34$\pm$0.22 & 0.85$\pm$0.02 & 0.44$\pm$0.28\\
        $t_{9}$ & 
        \textbf{0.53}$~\pm~$0.06 & \textbf{1.0}$~\pm~$0.01 & \textbf{0.69}$~\pm~$0.06 & 
        0.46$~\pm~$0.04 & 0.96$~\pm~$0.02 & 0.62$~\pm~$0.03 & 
        0.39$\pm$0.2 & 0.87$\pm$0.02 & 0.5$\pm$0.26\\
        $t_{10}$ & 
        \textbf{0.56}$~\pm~$0.06 & \textbf{1.0}$~\pm~$0.01 & \textbf{0.71}$~\pm~$0.05 & 
        0.46$~\pm~$0.03 & 0.96$~\pm~$0.02 & 0.62$~\pm~$0.03 & 
        0.38$\pm$0.2 & 0.86$\pm$0.02 & 0.49$\pm$0.26\\
% $t_{11}$ & 0.52$~\pm~$0.06 & 1.0$~\pm~$0.01 & 0.68$~\pm~$0.06 & 0.45$~\pm~$0.03 & 0.96$~\pm~$0.01 & 0.61$~\pm~$0.03 \\
    \bottomrule
    \end{tabular}
    }
    \label{tab:harmonic_custom}
\end{table}

\begin{table}[htb!]
    \centering
    \caption{\textsc{Dynamic} setting. TC stands for Topic Coherence, TD for Topic Diversity, and $H$ indicates the harmonic mean between Topic Coherence and Topic Diversity.}
    \resizebox{\columnwidth}{!}{
    \begin{tabular}{c|c|c|c||c|c|c||c|c|c}
    \toprule
    Time & \multicolumn{3}{c||}{\textsc{streamETM}} & \multicolumn{3}{c||}{\textsc{mergeBERT}} & \multicolumn{3}{c}{\textsc{onlineBERT}}\\
    \cmidrule{2-10}
         & TC & TD & H$_{(\text{TC,TD})}$ & TC & TD & H$_{(\text{TC,TD})}$ & TC & TD & H$_{(\text{TC,TD})}$\\
    \midrule
    $t_{0}$ & 
    \textbf{0.59}$~\pm~$0.10 & \textbf{0.98}$~\pm~$0.03 & \textbf{0.73}$~\pm~$0.08 & 
    0.56$~\pm~$0.05 & 0.96$~\pm~$0.02 & 0.71$~\pm~$0.04 & 
    0.49$\pm$0.03 & 0.81$\pm$0.02 & 0.61$\pm$0.02\\
    $t_{1}$ & 
    \textbf{0.58}$~\pm~$0.09 & \textbf{0.99}$~\pm~$0.02 & \textbf{0.73}$~\pm~$0.07 & 
    0.51$~\pm~$0.04 & 0.96$~\pm~$0.02 & 0.67$~\pm~$0.04 & 
    0.51$\pm$0.02 & 0.67$\pm$0.02 & 0.58$\pm$0.01\\
    $t_{2}$ & 
    \textbf{0.56}$~\pm~$0.06 & \textbf{0.98}$~\pm~$0.02 & \textbf{0.71}$~\pm~$0.05 & 
    0.49$~\pm~$0.03 & 0.96$~\pm~$0.02 & 0.65$~\pm~$0.03 & 
    0.5$\pm$0.02 & 0.8$\pm$0.02 & 0.61$\pm$0.02\\
    $t_{3}$ & 
    \textbf{0.62}$~\pm~$0.07 & \textbf{0.97}$~\pm~$0.03 & \textbf{0.75}$~\pm~$0.05 & 
    0.52$~\pm~$0.04 & 0.96$~\pm~$0.02 & 0.67$~\pm~$0.04 & 
    0.47$\pm$0.13 & 0.8$\pm$0.02 & 0.57$\pm$0.16\\
    $t_{4}$ & 
    \textbf{0.60}$ ~\pm~$0.04 & \textbf{0.97}$~\pm~$0.03 & \textbf{0.74}$~\pm~$0.04 & 
    0.49$~\pm~$0.04 & 0.95$~\pm~$0.02 & 0.65$~\pm~$0.04 & 
    0.51$\pm$0.03 & 0.81$\pm$0.02 & 0.62$\pm$0.03\\
    $t_{5}$ & 
    \textbf{0.59}$~\pm~$0.07 & \textbf{0.98}$~\pm~$0.03 & \textbf{0.73}$~\pm~$0.06 & 
    0.51$~\pm~$0.03 & 0.95$~\pm~$0.02 & 0.66$~\pm~$0.03 & 
    0.5$\pm$0.02 & 0.83$\pm$0.02 & 0.63$\pm$0.02\\
    $t_{6}$ & 
    \textbf{0.52}$~\pm~$0.17 & \textbf{0.97}$~\pm~$0.03 & \textbf{0.66}$~\pm~$0.19 & 
    0.45$~\pm~$0.03 & 0.95$~\pm~$0.02 & 0.61$~\pm~$0.03 & 
    0.37$\pm$0.23 & 0.83$\pm$0.01 & 0.46$\pm$0.29\\
    $t_{7}$ &
    \textbf{0.45}$~\pm~$0.25 & 0.95$~\pm~$0.04 & 0.57$~\pm~$0.30 & 
    0.44$~\pm~$0.03 & \textbf{0.96}$~\pm~$0.02 & \textbf{0.60}$ ~\pm~$0.03 & 
    0.33$\pm$0.24 & 0.82$\pm$0.02 & 0.41$\pm$0.3\\
    $t_{8}$ & 
    \textbf{0.51}$~\pm~$0.15 & 0.95$~\pm~$0.04 & \textbf{0.64}$~\pm~$0.19 & 
    0.46$~\pm~$0.03 & \textbf{0.96}$~\pm~$0.02 & 0.62$~\pm~$0.03 & 
    0.46$\pm$0.13 & 0.82$\pm$0.02 & 0.57$\pm$0.16\\
    $t_{9}$ & 
    \textbf{0.53}$~\pm~$0.06 & \textbf{0.95}$~\pm~$0.04 & \textbf{0.68}$~\pm~$0.05 & 
    0.45$~\pm~$0.04 & \textbf{0.95}$~\pm~$0.02 & 0.61$~\pm~$0.04 & 
    0.22$\pm$0.28 & 0.82$\pm$0.02 & 0.26$\pm$0.33\\
    $t_{10}$ & 
    \textbf{0.52}$~\pm~$0.06 & \textbf{0.96}$~\pm~$0.04 & \textbf{0.67}$~\pm~$0.05 & 
    0.46$~\pm~$0.03 & 0.95$~\pm~$0.02 & 0.61$~\pm~$0.03 & 
    0.34$\pm$0.29 & 0.83$\pm$0.02 & 0.41$\pm$0.34\\
    \bottomrule
    \end{tabular}
    }
    \label{tab:harmonic_extreme}
\end{table}

\subsection{Additional qualitative results}
\label{app:qualitative}
This section provides the top five words describing each topic across the 15 trainings for all the methods.
(cf.~\cref{tab:topics_custom,tab:topics_extreme}).
\begin{table}[htb!]
    \centering
    \caption{Qualitative assessment \textsc{Custom} setting. We present the top five words describing each topic across the 15 trainings. For \textsc{mergeBert}, we provide only five topics out of the approximately 23 identified by the model. We manually verify the topic associations to ensure consistency in topic indexing across the 15 trainings.}
    \resizebox{\columnwidth}{!}{
    \begin{tabular}{c|l|l|l}
    \toprule
    Time & \multicolumn{1}{c|}{\textsc{streamETM}} & \multicolumn{1}{c|}{\textsc{mergeBERT}} & \multicolumn{1}{c}{\textsc{onlineBERT}}\\
    \midrule
         $t_0$
         & 
         \textcolor{myblue}{wheel}, \textcolor{myblue}{manual}, \textcolor{myblue}{brake}, \textcolor{myblue}{tire}, \textcolor{myblue}{automatic}
         & \textcolor{myblue}{drive}, \textcolor{myblue}{engine}, \textcolor{myblue}{ford}, \textcolor{myblue}{dealer}, \textcolor{myblue}{price}
         & \textcolor{myblue}{dealer}, \textcolor{myblue}{price}, \textcolor{myblue}{odometer}, \textcolor{myblue}{model}, \textcolor{myblue}{discussion}\\
         & \textcolor{mygreen}{mission}, \textcolor{mygreen}{orbit}, \textcolor{mygreen}{spacecraft}, \textcolor{mygreen}{nasa}, \textcolor{mygreen}{shuttle}
         & \textcolor{mygreen}{space}, \textcolor{mygreen}{launch}, \textcolor{mygreen}{earth}, \textcolor{mygreen}{orbit}, \textcolor{mygreen}{mission}
         & \textcolor{mygreen}{pocket}, \textcolor{mygreen}{woman}, \textcolor{mygreen}{flat}, \textcolor{mygreen}{pant}, \textcolor{mygreen}{automotive}\\
         & 
        \textcolor{myorange}{jesus}, \textcolor{myorange}{christian}, \textcolor{myorange}{christians}, \textcolor{myorange}{bible}, \textcolor{myorange}{church}
        & \textcolor{myorange}{christian}, \textcolor{myorange}{people}, \textcolor{myorange}{religion}, \textcolor{myorange}{child}, \textcolor{myorange}{christians}
        & \textcolor{myorange}{line}, \textcolor{myorange}{magellan}, \textcolor{myorange}{jupiter}, \textcolor{myorange}{baalke}, \textcolor{myorange}{comet}\\
        
         \midrule
         $t_1$&
         \textcolor{myblue}{brake}, \textcolor{myblue}{wheel}, \textcolor{myblue}{manual}, \textcolor{myblue}{tire}, \textcolor{myblue}{automatic}
         & \textcolor{myblue}{drive}, \textcolor{myblue}{engine}, \textcolor{myblue}{ford}, \textcolor{myblue}{dealer}, \textcolor{myblue}{price}
         & \textcolor{myblue}{engine}, \textcolor{myblue}{price}, \textcolor{myblue}{dealer}, \textcolor{myblue}{mustang}, \textcolor{myblue}{line}\\
         & \textcolor{mygreen}{spacecraft}, \textcolor{mygreen}{mission}, \textcolor{mygreen}{orbit}, \textcolor{mygreen}{jupiter}, \textcolor{mygreen}{mars}
         & \textcolor{mygreen}{space}, \textcolor{mygreen}{launch}, \textcolor{mygreen}{earth}, \textcolor{mygreen}{orbit}, \textcolor{mygreen}{mission}
         & \textcolor{mygreen}{pocket}, \textcolor{mygreen}{woman}, \textcolor{mygreen}{pant}, \textcolor{mygreen}{flat}, \textcolor{mygreen}{wallet}\\
         &\textcolor{myorange}{christian}, \textcolor{myorange}{jesus}, \textcolor{myorange}{christians}, \textcolor{myorange}{bible}, \textcolor{myorange}{church}
         & \textcolor{myorange}{christian}, \textcolor{myorange}{people}, \textcolor{myorange}{religion}, \textcolor{myorange}{child}, \textcolor{myorange}{christians}
         & \textcolor{myorange}{line}, \textcolor{myorange}{subject}, \textcolor{myorange}{space}, \textcolor{myorange}{host}, \textcolor{myorange}{reply}\\
         \midrule
         
         $t_2$&
         \textcolor{myblue}{wheel}, \textcolor{myblue}{brake}, \textcolor{myblue}{tire}, \textcolor{myblue}{dealer}, \textcolor{myblue}{manual}
         & \textcolor{myblue}{drive}, \textcolor{myblue}{engine}, \textcolor{myblue}{ford}, \textcolor{myblue}{dealer}, \textcolor{myblue}{price}
         & \textcolor{myblue}{price}, \textcolor{myblue}{engine}, \textcolor{myblue}{model}, \textcolor{myblue}{saturn}, \textcolor{myblue}{indicator}\\
         & \textcolor{mygreen}{orbit}, \textcolor{mygreen}{spacecraft}, \textcolor{mygreen}{mission}, \textcolor{mygreen}{satellite}, \textcolor{mygreen}{jupiter}
         & \textcolor{mygreen}{space}, \textcolor{mygreen}{launch}, \textcolor{mygreen}{earth}, \textcolor{mygreen}{orbit}, \textcolor{mygreen}{mission}
         & \textcolor{mygreen}{dealer}, \textcolor{mygreen}{price}, \textcolor{mygreen}{honda}, \textcolor{mygreen}{unit}, \textcolor{mygreen}{subject}\\
         &\textcolor{myorange}{christian}, \textcolor{myorange}{christians}, \textcolor{myorange}{jesus}, \textcolor{myorange}{bible}, \textcolor{myorange}{religion}
         & \textcolor{myorange}{christian}, \textcolor{myorange}{people}, \textcolor{myorange}{religion}, \textcolor{myorange}{child}, \textcolor{myorange}{christians}
         & \textcolor{myorange}{write}, \textcolor{myorange}{jesus}, \textcolor{myorange}{religion}, \textcolor{myorange}{morality}, \textcolor{myorange}{christian}\\
         \midrule
         
         $t_3$&
         \textcolor{myblue}{wheel}, \textcolor{myblue}{driver}, \textcolor{myblue}{dealer}, \textcolor{myblue}{drive}, \textcolor{myblue}{engine}
         & \textcolor{myblue}{drive}, \textcolor{myblue}{engine}, \textcolor{myblue}{ford}, \textcolor{myblue}{dealer}, \textcolor{myblue}{price}
         & \textcolor{myblue}{drive}, \textcolor{myblue}{dealer}, \textcolor{myblue}{price}, \textcolor{myblue}{opel}, \textcolor{myblue}{indicator}\\
         & \textcolor{mygreen}{mission}, \textcolor{mygreen}{orbit}, \textcolor{mygreen}{spacecraft}, \textcolor{mygreen}{mars}, \textcolor{mygreen}{satellite}
         & \textcolor{mygreen}{space}, \textcolor{mygreen}{launch}, \textcolor{mygreen}{earth}, \textcolor{mygreen}{orbit}, \textcolor{mygreen}{mission}
         &\textcolor{mygreen}{integra}, \textcolor{mygreen}{dealer}, \textcolor{mygreen}{sedan}, \textcolor{mygreen}{import}, \textcolor{mygreen}{honda}\\
        &\textcolor{myorange}{christian}, \textcolor{myorange}{christians}, \textcolor{myorange}{jesus}, \textcolor{myorange}{bible}, \textcolor{myorange}{christ}
        & \textcolor{myorange}{christian}, \textcolor{myorange}{people}, \textcolor{myorange}{religion}, \textcolor{myorange}{child}, \textcolor{myorange}{christians}
        & \textcolor{myorange}{line}, \textcolor{myorange}{subject}, \textcolor{myorange}{write}, \textcolor{myorange}{space}, \textcolor{myorange}{film}\\
        \midrule
        
         $t_4$
         &
         \textcolor{myblue}{engine}, \textcolor{myblue}{wheel}, \textcolor{myblue}{driver}, \textcolor{myblue}{dealer}, \textcolor{myblue}{drive}
         & \textcolor{myblue}{drive}, \textcolor{myblue}{engine}, \textcolor{myblue}{ford}, \textcolor{myblue}{dealer}, \textcolor{myblue}{price}
         & \textcolor{myblue}{engine}, \textcolor{myblue}{fuel}, \textcolor{myblue}{like}, \textcolor{myblue}{plutonium}, \textcolor{myblue}{reprocess}\\
         & 
         \textcolor{mygreen}{mission}, \textcolor{mygreen}{orbit}, \textcolor{mygreen}{spacecraft}, \textcolor{mygreen}{satellite}, \textcolor{mygreen}{mars}
         & \textcolor{mygreen}{space}, \textcolor{mygreen}{launch}, \textcolor{mygreen}{earth}, \textcolor{mygreen}{orbit}, \textcolor{mygreen}{mission}
         & \textcolor{mygreen}{insurance}, \textcolor{mygreen}{houston}, \textcolor{mygreen}{rate}, \textcolor{mygreen}{info}, \textcolor{mygreen}{jmhhopper}\\
         &\textcolor{myorange}{christian}, \textcolor{myorange}{christians}, \textcolor{myorange}{bible}, \textcolor{myorange}{jesus}, \textcolor{myorange}{christ}
         & \textcolor{myorange}{christian}, \textcolor{myorange}{people}, \textcolor{myorange}{religion}, \textcolor{myorange}{child}, \textcolor{myorange}{christians}
         & \textcolor{myorange}{line}, \textcolor{myorange}{subject}, \textcolor{myorange}{kent}, \textcolor{myorange}{host}, \textcolor{myorange}{solar}\\
         
         \midrule
         $t_5$&\textcolor{myblue}{ford}, \textcolor{myblue}{engine}, \textcolor{myblue}{driver}, \textcolor{myblue}{wheel}, \textcolor{myblue}{dealer}
         & \textcolor{myblue}{drive}, \textcolor{myblue}{engine}, \textcolor{myblue}{ford}, \textcolor{myblue}{dealer}, \textcolor{myblue}{price}
         & \textcolor{myblue}{mustang}, \textcolor{myblue}{radar}, \textcolor{myblue}{ford}, \textcolor{myblue}{brake}, \textcolor{myblue}{tire}\\
         &
         \textcolor{mygreen}{mission}, \textcolor{mygreen}{spacecraft}, \textcolor{mygreen}{orbit}, \textcolor{mygreen}{mars}, \textcolor{mygreen}{satellite}
         & \textcolor{mygreen}{space}, \textcolor{mygreen}{launch}, \textcolor{mygreen}{earth}, \textcolor{mygreen}{orbit}, \textcolor{mygreen}{mission}
         & \textcolor{mygreen}{insurance}, \textcolor{mygreen}{odometer}, \textcolor{mygreen}{owner}, \textcolor{mygreen}{driver}, \textcolor{mygreen}{dealer}\\
         &\textcolor{myorange}{christian}, \textcolor{myorange}{jesus}, \textcolor{myorange}{bible}, \textcolor{myorange}{christians}, \textcolor{myorange}{christ}
         & \textcolor{myorange}{christian}, \textcolor{myorange}{people}, \textcolor{myorange}{religion}, \textcolor{myorange}{child}, \textcolor{myorange}{christians}
         & \textcolor{myorange}{write}, \textcolor{myorange}{subject}, \textcolor{myorange}{line}, \textcolor{myorange}{moon}, \textcolor{myorange}{university}\\
         \midrule
         
         $t_6$&\textcolor{myblue}{engine}, \textcolor{myblue}{driver}, \textcolor{myblue}{ford}, \textcolor{myblue}{wheel}, \textcolor{myblue}{dealer}
         & \textcolor{myblue}{drive}, \textcolor{myblue}{engine}, \textcolor{myblue}{ford}, \textcolor{myblue}{dealer}, \textcolor{myblue}{price}
         & \textcolor{myblue}{mustang}, \textcolor{myblue}{engine}, \textcolor{myblue}{clutch}, \textcolor{myblue}{fuel}, \textcolor{myblue}{dealer}
         \\
         &
         \textcolor{mygreen}{space}, \textcolor{mygreen}{launch}, \textcolor{mygreen}{rocket}, \textcolor{mygreen}{mission}, \textcolor{mygreen}{mars}
         & \textcolor{mygreen}{space}, \textcolor{mygreen}{launch}, \textcolor{mygreen}{earth}, \textcolor{mygreen}{orbit}, \textcolor{mygreen}{mission}
         & \textcolor{mygreen}{villager}, \textcolor{mygreen}{saturn}, \textcolor{mygreen}{sacramento}, \textcolor{mygreen}{mihir}, \textcolor{mygreen}{geof}\\
         &\textcolor{myorange}{christian}, \textcolor{myorange}{jesus}, \textcolor{myorange}{christ}, \textcolor{myorange}{christians}, \textcolor{myorange}{bible}
         & \textcolor{myorange}{christian}, \textcolor{myorange}{people}, \textcolor{myorange}{religion}, \textcolor{myorange}{child}, \textcolor{myorange}{christians}
         & \textcolor{myorange}{jesus}, \textcolor{myorange}{kent}, \textcolor{myorange}{article}, \textcolor{myorange}{turkey}, \textcolor{myorange}{jews}\\
         && \textcolor{violet}{jewish}, \textcolor{violet}{baseball}, \textcolor{violet}{players}, \textcolor{violet}{long}, \textcolor{violet}{list}\\
         \midrule
         
         $t_7$&\textcolor{myblue}{engine}, \textcolor{myblue}{ford}, \textcolor{myblue}{wheel}, \textcolor{myblue}{driver}, \textcolor{myblue}{dealer}
         & \textcolor{myblue}{drive}, \textcolor{myblue}{engine}, \textcolor{myblue}{ford}, \textcolor{myblue}{dealer}, \textcolor{myblue}{price}
         & \textcolor{myblue}{engine}, \textcolor{myblue}{mustang}, \textcolor{myblue}{ford}, \textcolor{myblue}{diesel}, \textcolor{myblue}{model}\\
         &
         \textcolor{mygreen}{field}, \textcolor{mygreen}{center}, \textcolor{mygreen}{space}, \textcolor{mygreen}{rocket}, \textcolor{mygreen}{launch}
         & \textcolor{mygreen}{space}, \textcolor{mygreen}{launch}, \textcolor{mygreen}{earth}, \textcolor{mygreen}{orbit}, \textcolor{mygreen}{mission}
         & \textcolor{mygreen}{game}, \textcolor{mygreen}{pitcher}, \textcolor{mygreen}{line}, \textcolor{mygreen}{field}, \textcolor{mygreen}{white}\\
         &\textcolor{myorange}{christian}, \textcolor{myorange}{jesus}, \textcolor{myorange}{christians}, \textcolor{myorange}{christ}, \textcolor{myorange}{bible}
        & \textcolor{myorange}{christian}, \textcolor{myorange}{people}, \textcolor{myorange}{religion}, \textcolor{myorange}{child}, \textcolor{myorange}{christians}
        & \textcolor{myorange}{christian}, \textcolor{myorange}{jesus}, \textcolor{myorange}{people}, \textcolor{myorange}{word}, \textcolor{myorange}{bible}\\
         && \textcolor{violet}{jewish}, \textcolor{violet}{baseball}, \textcolor{violet}{game}, \textcolor{violet}{players}, \textcolor{violet}{humor}
         \\
         
         \midrule
         $t_8$&\textcolor{myblue}{engine}, \textcolor{myblue}{ford}, \textcolor{myblue}{wheel}, \textcolor{myblue}{driver}, \textcolor{myblue}{dealer}
         & \textcolor{myblue}{drive}, \textcolor{myblue}{engine}, \textcolor{myblue}{ford}, \textcolor{myblue}{dealer}, \textcolor{myblue}{price}
         & \textcolor{myblue}{brake}, \textcolor{myblue}{price}, \textcolor{myblue}{engine}, \textcolor{myblue}{honda}, \textcolor{myblue}{diesel}\\
         & 
         \textcolor{mygreen}{center}, \textcolor{mygreen}{space}, \textcolor{mygreen}{field}, \textcolor{mygreen}{star}, \textcolor{mygreen}{system}
         & \textcolor{mygreen}{space}, \textcolor{mygreen}{launch}, \textcolor{mygreen}{earth}, \textcolor{mygreen}{orbit}, \textcolor{mygreen}{mission}
         &\textcolor{mygreen}{train}, \textcolor{mygreen}{rain}, \textcolor{mygreen}{pathfinder}, \textcolor{mygreen}{thatch}, \textcolor{mygreen}{stevens}\\
         &\textcolor{myorange}{christian}, \textcolor{myorange}{jesus}, \textcolor{myorange}{christ}, \textcolor{myorange}{christians}, \textcolor{myorange}{bible}
        & \textcolor{myorange}{christian}, \textcolor{myorange}{people}, \textcolor{myorange}{religion}, \textcolor{myorange}{child}, \textcolor{myorange}{christians}
        & \textcolor{myorange}{jesus}, \textcolor{myorange}{people}, \textcolor{myorange}{silence}, \textcolor{myorange}{write}, \textcolor{myorange}{article}\\
        && \textcolor{violet}{jewish}, \textcolor{violet}{baseball}, \textcolor{violet}{players}, \textcolor{violet}{mets}, \textcolor{violet}{game}\\
        &&& \textcolor{darkred}{brian}, \textcolor{darkred}{radar}, \textcolor{darkred}{detector}, \textcolor{darkred}{light}, \textcolor{darkred}{conceal}\\
         \midrule
         
         $t_{9}$ &
        \textcolor{myblue}{engine}, \textcolor{myblue}{ford}, \textcolor{myblue}{wheel}, \textcolor{myblue}{driver}, \textcolor{myblue}{dealer}
        & \textcolor{myblue}{drive}, \textcolor{myblue}{engine}, \textcolor{myblue}{ford}, \textcolor{myblue}{dealer}, \textcolor{myblue}{price}
        & \textcolor{myblue}{engine}, \textcolor{myblue}{fuel}, \textcolor{myblue}{drive}, \textcolor{myblue}{toyota}, \textcolor{myblue}{seat}\\
        & 
        \textcolor{mygreen}{center}, \textcolor{mygreen}{field}, \textcolor{mygreen}{space}, \textcolor{mygreen}{team}, \textcolor{mygreen}{star}
        & \textcolor{mygreen}{space}, \textcolor{mygreen}{launch}, \textcolor{mygreen}{earth}, \textcolor{mygreen}{orbit}, \textcolor{mygreen}{mission}
        & \textcolor{mygreen}{smith}, \textcolor{mygreen}{list}, \textcolor{mygreen}{ticket}, \textcolor{mygreen}{lewis}, \textcolor{mygreen}{license}\\
        &\textcolor{myorange}{jesus}, \textcolor{myorange}{bible}, \textcolor{myorange}{christ}, \textcolor{myorange}{christian}, \textcolor{myorange}{faith}
        &\textcolor{myorange}{christian}, \textcolor{myorange}{people}, \textcolor{myorange}{religion}, \textcolor{myorange}{child}, \textcolor{myorange}{christians}
         & \textcolor{myorange}{christian}, \textcolor{myorange}{mistake}, \textcolor{myorange}{silence}, \textcolor{myorange}{moment}, \textcolor{myorange}{jesus}\\
        && \textcolor{violet}{jewish}, \textcolor{violet}{baseball}, \textcolor{violet}{players}, \textcolor{violet}{humor}, \textcolor{violet}{season}\\
        && \textcolor{darkred}{cancer}, \textcolor{darkred}{vitamin}, \textcolor{darkred}{stone}, \textcolor{darkred}{intake}, \textcolor{darkred}{medical}
        & \textcolor{darkred}{food}, \textcolor{darkred}{vasectomy}, \textcolor{darkred}{address}, \textcolor{darkred}{cancer}, \textcolor{darkred}{doctor}\\
        \midrule
        
         $t_{10}$ &
        \textcolor{myblue}{engine}, \textcolor{myblue}{ford}, \textcolor{myblue}{wheel}, \textcolor{myblue}{driver}, \textcolor{myblue}{drive}
        & \textcolor{myblue}{drive}, \textcolor{myblue}{engine}, \textcolor{myblue}{ford}, \textcolor{myblue}{dealer}, \textcolor{myblue}{price}
        & \textcolor{myblue}{engine}, \textcolor{myblue}{drive}, \textcolor{myblue}{speed}, \textcolor{myblue}{line}, \textcolor{myblue}{subject}\\
        & 
        \textcolor{mygreen}{center}, \textcolor{mygreen}{field}, \textcolor{mygreen}{team}, \textcolor{mygreen}{program}, \textcolor{mygreen}{star}
        & \textcolor{mygreen}{space}, \textcolor{mygreen}{launch}, \textcolor{mygreen}{earth}, \textcolor{mygreen}{orbit}, \textcolor{mygreen}{mission}
        & \textcolor{mygreen}{talon}, \textcolor{mygreen}{plate}, \textcolor{mygreen}{info}, \textcolor{mygreen}{license}, \textcolor{mygreen}{ticket}\\
        &\textcolor{myorange}{jesus}, \textcolor{myorange}{bible}, \textcolor{myorange}{christ}, \textcolor{myorange}{christian}, \textcolor{myorange}{faith}
        &\textcolor{myorange}{christian}, \textcolor{myorange}{people}, \textcolor{myorange}{religion}, \textcolor{myorange}{child}, \textcolor{myorange}{christians}
        &\textcolor{myorange}{jesus}, \textcolor{myorange}{christian}, \textcolor{myorange}{bible}, \textcolor{myorange}{matthew}, \textcolor{myorange}{morality}\\
        && \textcolor{violet}{jewish}, \textcolor{violet}{baseball}, \textcolor{violet}{players}, \textcolor{violet}{humor}, \textcolor{violet}{season}
        & \textcolor{violet}{game}, \textcolor{violet}{team}, \textcolor{violet}{uniform}, \textcolor{violet}{runner}, \textcolor{violet}{dominance}\\
        &&\textcolor{darkred}{cancer}, \textcolor{darkred}{vitamin}, \textcolor{darkred}{stone}, \textcolor{darkred}{intake}, \textcolor{darkred}{medical}
        & \textcolor{darkred}{real}, \textcolor{darkred}{food}, \textcolor{darkred}{john}, \textcolor{darkred}{influence}, \textcolor{darkred}{nielsen}\\
        \midrule
        
         $t_{11}$ &
        \textcolor{myblue}{engine}, \textcolor{myblue}{ford}, \textcolor{myblue}{wheel}, \textcolor{myblue}{driver}, \textcolor{myblue}{drive}
        & \textcolor{myblue}{drive}, \textcolor{myblue}{engine}, \textcolor{myblue}{ford}, \textcolor{myblue}{dealer}, \textcolor{myblue}{price}
        & \textcolor{myblue}{engine}, \textcolor{myblue}{brake}, \textcolor{myblue}{tire}, \textcolor{myblue}{fuel}, \textcolor{myblue}{ford}\\
        & 
        \textcolor{mygreen}{center}, \textcolor{mygreen}{field}, \textcolor{mygreen}{team}, \textcolor{mygreen}{manager}, \textcolor{mygreen}{first}
        & \textcolor{mygreen}{space}, \textcolor{mygreen}{launch}, \textcolor{mygreen}{earth}, \textcolor{mygreen}{orbit}, \textcolor{mygreen}{mission}
        & \textcolor{mygreen}{horizon}, \textcolor{mygreen}{black}, \textcolor{mygreen}{event}, \textcolor{mygreen}{zeno}, \textcolor{mygreen}{paradox}\\
        &\textcolor{myorange}{jesus}, \textcolor{myorange}{christ}, \textcolor{myorange}{christian}, \textcolor{myorange}{bible}, \textcolor{myorange}{religion}
        &\textcolor{myorange}{christian}, \textcolor{myorange}{people}, \textcolor{myorange}{religion}, \textcolor{myorange}{child}, \textcolor{myorange}{christians}
        & \textcolor{myorange}{jesus}, \textcolor{myorange}{bible}, \textcolor{myorange}{christian}, \textcolor{myorange}{religion}, \textcolor{myorange}{people}\\
        && \textcolor{violet}{jewish}, \textcolor{violet}{baseball}, \textcolor{violet}{players}, \textcolor{violet}{humor}, \textcolor{violet}{season}
        & \textcolor{violet}{game}, \textcolor{violet}{baseball}, \textcolor{violet}{team}, \textcolor{violet}{play}, \textcolor{violet}{year}\\
        &&\textcolor{darkred}{cancer}, \textcolor{darkred}{medical}, \textcolor{darkred}{shot}, \textcolor{darkred}{tumor}, \textcolor{darkred}{center}
        & \textcolor{darkred}{diet}, \textcolor{darkred}{placebo}, \textcolor{darkred}{subject}, \textcolor{darkred}{line}, \textcolor{darkred}{yeast}
        \\
    \bottomrule
    \end{tabular}
    }
    \label{tab:topics_custom}
\end{table}
\begin{table}[htb!]
    \centering
    \caption{Qualitative assessment \textsc{Dynamic} setting. We present the top five words describing each topic across the 15 trainings. For \textsc{mergeBERT}, we provide only five topics out of the approximately 23 identified by the model. We manually verify the topic associations to ensure consistency in topic indexing across the 15 trainings.}
    \resizebox{\columnwidth}{!}{
    \begin{tabular}{c|l|l|l}
    \toprule
    Time & \multicolumn{1}{c|}{\textsc{streamETM}} & \multicolumn{1}{c|}{\textsc{mergeBERT}} &  \multicolumn{1}{c}{\textsc{onlineBERT}}\\
    \midrule
         $t_0$ & 
         \textcolor{darkblue}{sale}, \textcolor{darkblue}{manual}, \textcolor{darkblue}{disk}, \textcolor{darkblue}{card}, \textcolor{darkblue}{printer}
         && \\
         & \textcolor{darkgreen}{security}, \textcolor{darkgreen}{chip}, \textcolor{darkgreen}{algorithm}, \textcolor{darkgreen}{device}, \textcolor{darkgreen}{privacy}
         & \textcolor{darkgreen}{clipper}, \textcolor{darkgreen}{government}, \textcolor{darkgreen}{encryption}, \textcolor{darkgreen}{chip}, \textcolor{darkgreen}{algorithm}
         & \textcolor{darkgreen}{government}, \textcolor{darkgreen}{encryption}, \textcolor{darkgreen}{clinton}, \textcolor{darkgreen}{believe}, \textcolor{darkgreen}{mary}\\
         & \textcolor{mediumorchid}{jesus}, \textcolor{mediumorchid}{christ}, \textcolor{mediumorchid}{marriage}, \textcolor{mediumorchid}{faith}, \textcolor{mediumorchid}{spirit}
         & \textcolor{mediumorchid}{church}, \textcolor{mediumorchid}{book}, \textcolor{mediumorchid}{christian}, \textcolor{mediumorchid}{life}, \textcolor{mediumorchid}{faith}
         & \textcolor{mediumorchid}{clipper}, \textcolor{mediumorchid}{mary}, \textcolor{mediumorchid}{assumption}, \textcolor{mediumorchid}{voice}, \textcolor{mediumorchid}{intercon}\\
         &&&\\
         &&\textcolor{red}{sale}, \textcolor{red}{sell}, \textcolor{red}{nntp}, \textcolor{red}{offer}, \textcolor{red}{host}
         & \textcolor{red}{sale}, \textcolor{red}{nntp}, \textcolor{red}{host}, \textcolor{red}{posting}, \textcolor{red}{price}\\
        \midrule
         
         $t_{1}$ & \textcolor{darkblue}{disk}, \textcolor{darkblue}{card}, \textcolor{darkblue}{sale}, \textcolor{darkblue}{software}, \textcolor{darkblue}{printer} 
         &&\\
         & \textcolor{darkgreen}{security}, \textcolor{darkgreen}{device}, \textcolor{darkgreen}{chip}, \textcolor{darkgreen}{algorithm}, \textcolor{darkgreen}{digital}
         & \textcolor{darkgreen}{encryption}, \textcolor{darkgreen}{clipper}, \textcolor{darkgreen}{government}, \textcolor{darkgreen}{chip}, \textcolor{darkgreen}{algorithm}
         & \textcolor{darkgreen}{government}, \textcolor{darkgreen}{homosexual}, \textcolor{darkgreen}{subject}, \textcolor{darkgreen}{clipper}, \textcolor{darkgreen}{encryption}\\
         & \textcolor{mediumorchid}{jesus}, \textcolor{mediumorchid}{christ}, \textcolor{mediumorchid}{marriage}, \textcolor{mediumorchid}{spirit}, \textcolor{mediumorchid}{faith}
         & \textcolor{mediumorchid}{church}, \textcolor{mediumorchid}{book}, \textcolor{mediumorchid}{christian}, \textcolor{mediumorchid}{life}, \textcolor{mediumorchid}{faith}
         & \textcolor{mediumorchid}{line}, \textcolor{mediumorchid}{sale}, \textcolor{mediumorchid}{nntp}, \textcolor{mediumorchid}{subject}, \textcolor{mediumorchid}{clipper}\\
         &&&\\
         &&\textcolor{red}{sale}, \textcolor{red}{sell}, \textcolor{red}{nntp}, \textcolor{red}{offer}, \textcolor{red}{host}
         & \textcolor{red}{subject}, \textcolor{red}{sale}, \textcolor{red}{line}, \textcolor{red}{nntp}, \textcolor{red}{ticket}\\
         \midrule
         
         $t_{2}$ & \textcolor{darkblue}{sale}, \textcolor{darkblue}{card}, \textcolor{darkblue}{printer}, \textcolor{darkblue}{drive}, \textcolor{darkblue}{video} &&\\
         & \textcolor{darkgreen}{security}, \textcolor{darkgreen}{telephone}, \textcolor{darkgreen}{device}, \textcolor{darkgreen}{chip}, \textcolor{darkgreen}{software}
         & \textcolor{darkgreen}{encryption}, \textcolor{darkgreen}{clipper}, \textcolor{darkgreen}{government}, \textcolor{darkgreen}{chip}, \textcolor{darkgreen}{algorithm}
         & \textcolor{darkgreen}{encryption}, \textcolor{darkgreen}{chip}, \textcolor{darkgreen}{clipper}, \textcolor{darkgreen}{government}, \textcolor{darkgreen}{people}\\
         & \textcolor{mediumorchid}{jesus}, \textcolor{mediumorchid}{spirit}, \textcolor{mediumorchid}{christ}, \textcolor{mediumorchid}{marriage}, \textcolor{mediumorchid}{faith}
         & \textcolor{mediumorchid}{church}, \textcolor{mediumorchid}{book}, \textcolor{mediumorchid}{christian}, \textcolor{mediumorchid}{life}, \textcolor{mediumorchid}{faith}
         & \textcolor{mediumorchid}{chip}, \textcolor{mediumorchid}{sale}, \textcolor{mediumorchid}{game}, \textcolor{mediumorchid}{clipper}, \textcolor{mediumorchid}{bible}\\
         &&&\\
         &&\textcolor{red}{sale}, \textcolor{red}{sell}, \textcolor{red}{nntp}, \textcolor{red}{offer}, \textcolor{red}{host}
         & \textcolor{red}{sale}, \textcolor{red}{host}, \textcolor{red}{subject}, \textcolor{red}{line}, \textcolor{red}{ticket}\\
         \midrule
         
         $t_{3}$ & \textcolor{darkblue}{software}, \textcolor{darkblue}{video}, \textcolor{darkblue}{sale}, \textcolor{darkblue}{card}, \textcolor{darkblue}{printer}
         &\textcolor{darkblue}{printer}, \textcolor{darkblue}{compatible}, \textcolor{darkblue}{driver}, \textcolor{darkblue}{brand}, \textcolor{darkblue}{stamp}\\
         & \textcolor{darkgreen}{telephone}, \textcolor{darkgreen}{security}, \textcolor{darkgreen}{system}, \textcolor{darkgreen}{device}, \textcolor{darkgreen}{software}
         & \textcolor{darkgreen}{encryption}, \textcolor{darkgreen}{clipper}, \textcolor{darkgreen}{government}, \textcolor{darkgreen}{chip}, \textcolor{darkgreen}{algorithm}
         & \textcolor{darkgreen}{clipper}, \textcolor{darkgreen}{government}, \textcolor{darkgreen}{encryption}, \textcolor{darkgreen}{chip}, \textcolor{darkgreen}{love}\\
         & \textcolor{mediumorchid}{jesus}, \textcolor{mediumorchid}{faith}, \textcolor{mediumorchid}{christ}, \textcolor{mediumorchid}{marriage}, \textcolor{mediumorchid}{christians}
         & \textcolor{mediumorchid}{church}, \textcolor{mediumorchid}{book}, \textcolor{mediumorchid}{christian}, \textcolor{mediumorchid}{life}, \textcolor{mediumorchid}{faith}
         & \textcolor{mediumorchid}{line}, \textcolor{mediumorchid}{hate}, \textcolor{mediumorchid}{love}, \textcolor{mediumorchid}{faith}, \textcolor{mediumorchid}{sale}\\
         & \textcolor{orange}{armenians}, \textcolor{orange}{turkish}, \textcolor{orange}{turkey}, \textcolor{orange}{armenian}, \textcolor{orange}{armenia}
         & \textcolor{orange}{soviet}, \textcolor{orange}{greek}, \textcolor{orange}{armenian}, \textcolor{orange}{turks}, \textcolor{orange}{turkish}\\
         &&\textcolor{red}{sale}, \textcolor{red}{sell}, \textcolor{red}{nntp}, \textcolor{red}{offer}, \textcolor{red}{host}
         & \textcolor{red}{sale}, \textcolor{red}{subject}, \textcolor{red}{line}, \textcolor{red}{ticket}, \textcolor{red}{nntp}\\
         \midrule
         
        $t_{4}$ & \textcolor{darkblue}{software}, \textcolor{darkblue}{video}, \textcolor{darkblue}{card}, \textcolor{darkblue}{printer}, \textcolor{darkblue}{disk}
        & \textcolor{darkblue}{printer}, \textcolor{darkblue}{compatible}, \textcolor{darkblue}{driver}, \textcolor{darkblue}{brand}, \textcolor{darkblue}{stamp}\\
        & \textcolor{darkgreen}{system}, \textcolor{darkgreen}{telephone}, \textcolor{darkgreen}{security}, \textcolor{darkgreen}{phone}, \textcolor{darkgreen}{software}
        & \textcolor{darkgreen}{encryption}, \textcolor{darkgreen}{clipper}, \textcolor{darkgreen}{government}, \textcolor{darkgreen}{chip}, \textcolor{darkgreen}{algorithm}
        & \textcolor{darkgreen}{clipper}, \textcolor{darkgreen}{encryption}, \textcolor{darkgreen}{government}, \textcolor{darkgreen}{chip}, \textcolor{darkgreen}{israel}\\
        & \textcolor{mediumorchid}{faith}, \textcolor{mediumorchid}{jesus}, \textcolor{mediumorchid}{christ}, \textcolor{mediumorchid}{christians}, \textcolor{mediumorchid}{marriage}
        & \textcolor{mediumorchid}{church}, \textcolor{mediumorchid}{book}, \textcolor{mediumorchid}{christian}, \textcolor{mediumorchid}{life}, \textcolor{mediumorchid}{faith}
        & \textcolor{mediumorchid}{jesus}, \textcolor{mediumorchid}{chip}, \textcolor{mediumorchid}{christ}, \textcolor{mediumorchid}{truth}, \textcolor{mediumorchid}{believe}\\
        & \textcolor{orange}{turkish}, \textcolor{orange}{armenia}, \textcolor{orange}{turkey}, \textcolor{orange}{turks}, \textcolor{orange}{armenians}
        & \textcolor{orange}{soviet}, \textcolor{orange}{greek}, \textcolor{orange}{armenian}, \textcolor{orange}{turks}, \textcolor{orange}{turkish}\\
        &&\textcolor{red}{sale}, \textcolor{red}{sell}, \textcolor{red}{nntp}, \textcolor{red}{offer}, \textcolor{red}{host}
        & \textcolor{red}{sale}, \textcolor{red}{offer}, \textcolor{red}{tape}, \textcolor{red}{manual}, \textcolor{red}{receiver}\\
        \midrule
        
        $t_{5}$ & \textcolor{darkblue}{software}, \textcolor{darkblue}{video}, \textcolor{darkblue}{computer}, \textcolor{darkblue}{card}, \textcolor{darkblue}{disk}
        & \textcolor{darkblue}{printer}, \textcolor{darkblue}{compatible}, \textcolor{darkblue}{driver}, \textcolor{darkblue}{brand}, \textcolor{darkblue}{stamp}\\
        & \textcolor{darkgreen}{system}, \textcolor{darkgreen}{security}, \textcolor{darkgreen}{computer}, \textcolor{darkgreen}{phone}, \textcolor{darkgreen}{digital}
        & \textcolor{darkgreen}{encryption}, \textcolor{darkgreen}{clipper}, \textcolor{darkgreen}{government}, \textcolor{darkgreen}{chip}, \textcolor{darkgreen}{algorithm}
        & \textcolor{darkgreen}{encryption}, \textcolor{darkgreen}{clipper}, \textcolor{darkgreen}{chip}, \textcolor{darkgreen}{government}, \textcolor{darkgreen}{escrow}\\
        & \textcolor{mediumorchid}{faith}, \textcolor{mediumorchid}{jesus}, \textcolor{mediumorchid}{christ}, \textcolor{mediumorchid}{christians}, \textcolor{mediumorchid}{marriage}
        & \textcolor{mediumorchid}{church}, \textcolor{mediumorchid}{book}, \textcolor{mediumorchid}{christian}, \textcolor{mediumorchid}{life}, \textcolor{mediumorchid}{faith}
        & \textcolor{mediumorchid}{jesus}, \textcolor{mediumorchid}{church}, \textcolor{mediumorchid}{scripture}, \textcolor{mediumorchid}{christ}, \textcolor{mediumorchid}{believe}\\
        & \textcolor{orange}{turkish}, \textcolor{orange}{turkey}, \textcolor{orange}{turks}, \textcolor{orange}{armenians}, \textcolor{orange}{armenian}
        & \textcolor{orange}{soviet}, \textcolor{orange}{greek}, \textcolor{orange}{armenian}, \textcolor{orange}{turks}, \textcolor{orange}{turkish}\\
        &&\textcolor{red}{sale}, \textcolor{red}{sell}, \textcolor{red}{nntp}, \textcolor{red}{offer}, \textcolor{red}{host} 
        & \textcolor{red}{sale}, \textcolor{red}{line}, \textcolor{red}{subject}, \textcolor{red}{offer}, \textcolor{red}{tape}\\
        \midrule
        
        $t_{6}$ & \textcolor{darkblue}{software}, \textcolor{darkblue}{computer}, \textcolor{darkblue}{video}, \textcolor{darkblue}{printer}, \textcolor{darkblue}{card}
        & \textcolor{darkblue}{disk}, \textcolor{darkblue}{game}, \textcolor{darkblue}{snes}, \textcolor{darkblue}{genesis}, \textcolor{darkblue}{games}
        & \textcolor{darkblue}{sale}, \textcolor{darkblue}{drive}, \textcolor{darkblue}{offer}, \textcolor{darkblue}{movement}, \textcolor{darkblue}{church}\\
        & \textcolor{darkgreen}{system}, \textcolor{darkgreen}{computer}, \textcolor{darkgreen}{security}, \textcolor{darkgreen}{user}, \textcolor{darkgreen}{software}
        & \textcolor{darkgreen}{encryption}, \textcolor{darkgreen}{clipper}, \textcolor{darkgreen}{government}, \textcolor{darkgreen}{chip}, \textcolor{darkgreen}{algorithm}
        & \textcolor{darkgreen}{encryption}, \textcolor{darkgreen}{government}, \textcolor{darkgreen}{chip}, \textcolor{darkgreen}{clipper}, \textcolor{darkgreen}{security}\\
        & \textcolor{mediumorchid}{faith}, \textcolor{mediumorchid}{christ}, \textcolor{mediumorchid}{jesus}, \textcolor{mediumorchid}{christians}, \textcolor{mediumorchid}{church}
        & \textcolor{mediumorchid}{church}, \textcolor{mediumorchid}{book}, \textcolor{mediumorchid}{christian}, \textcolor{mediumorchid}{life}, \textcolor{mediumorchid}{faith}
        & \textcolor{mediumorchid}{clipper}, \textcolor{mediumorchid}{chip}, \textcolor{mediumorchid}{christians}, \textcolor{mediumorchid}{exist}, \textcolor{mediumorchid}{sale}\\
        & \textcolor{orange}{turkish}, \textcolor{orange}{turkey}, \textcolor{orange}{turks}, \textcolor{orange}{armenians}, \textcolor{orange}{armenian}
        & \textcolor{orange}{soviet}, \textcolor{orange}{greek}, \textcolor{orange}{armenian}, \textcolor{orange}{turks}, \textcolor{orange}{turkish}\\
        &&\textcolor{red}{sale}, \textcolor{red}{sell}, \textcolor{red}{nntp}, \textcolor{red}{offer}, \textcolor{red}{host}
        & \textcolor{red}{sale}, \textcolor{red}{line}, \textcolor{red}{offer}, \textcolor{red}{channel}, \textcolor{red}{subject}\\
        \midrule
        
        $t_{7}$ & \textcolor{darkblue}{software}, \textcolor{darkblue}{computer}, \textcolor{darkblue}{video}, \textcolor{darkblue}{disk}, \textcolor{darkblue}{card}
        & \textcolor{darkblue}{disk}, \textcolor{darkblue}{game}, \textcolor{darkblue}{snes}, \textcolor{darkblue}{genesis}, \textcolor{darkblue}{games}
        & \textcolor{darkblue}{scsi}, \textcolor{darkblue}{offer}, \textcolor{darkblue}{mary}, \textcolor{darkblue}{drive}, \textcolor{darkblue}{disk}\\
        & \textcolor{darkgreen}{system}, \textcolor{darkgreen}{computer}, \textcolor{darkgreen}{software}, \textcolor{darkgreen}{access}, \textcolor{darkgreen}{digital}
        & \textcolor{darkgreen}{encryption}, \textcolor{darkgreen}{clipper}, \textcolor{darkgreen}{government}, \textcolor{darkgreen}{chip}, \textcolor{darkgreen}{algorithm}
        & \textcolor{darkgreen}{encryption}, \textcolor{darkgreen}{clipper}, \textcolor{darkgreen}{chip}, \textcolor{darkgreen}{government}, \textcolor{darkgreen}{people}\\
        & \textcolor{mediumorchid}{faith}, \textcolor{mediumorchid}{christians}, \textcolor{mediumorchid}{church}, \textcolor{mediumorchid}{christ}, \textcolor{mediumorchid}{jesus}
        & \textcolor{mediumorchid}{church}, \textcolor{mediumorchid}{book}, \textcolor{mediumorchid}{christian}, \textcolor{mediumorchid}{life}, \textcolor{mediumorchid}{faith}
        & \textcolor{mediumorchid}{jesus}, \textcolor{mediumorchid}{subject}, \textcolor{mediumorchid}{sale}, \textcolor{mediumorchid}{line}, \textcolor{mediumorchid}{hell}\\
        & \textcolor{orange}{turkish}, \textcolor{orange}{turkey}, \textcolor{orange}{turks}, \textcolor{orange}{armenians}, \textcolor{orange}{armenian}
        & \textcolor{orange}{soviet}, \textcolor{orange}{greek}, \textcolor{orange}{armenian}, \textcolor{orange}{turks}, \textcolor{orange}{turkish}\\
        &&\textcolor{red}{sale}, \textcolor{red}{sell}, \textcolor{red}{nntp}, \textcolor{red}{offer}, \textcolor{red}{host}
        & \textcolor{red}{sale}, \textcolor{red}{offer}, \textcolor{red}{line}, \textcolor{red}{price}, \textcolor{red}{university}\\
        \midrule
        
        $t_{8}$ & \textcolor{darkblue}{software}, \textcolor{darkblue}{computer}, \textcolor{darkblue}{video}, \textcolor{darkblue}{system}, \textcolor{darkblue}{disk}
        & \textcolor{darkblue}{disk}, \textcolor{darkblue}{game}, \textcolor{darkblue}{snes}, \textcolor{darkblue}{genesis}, \textcolor{darkblue}{games}
        & \textcolor{darkblue}{drive}, \textcolor{darkblue}{card}, \textcolor{darkblue}{christian}, \textcolor{darkblue}{monitor}, \textcolor{darkblue}{vote}\\
        & \textcolor{darkgreen}{computer}, \textcolor{darkgreen}{system}, \textcolor{darkgreen}{software}, \textcolor{darkgreen}{access}, \textcolor{darkgreen}{security}
        & \textcolor{darkgreen}{encryption}, \textcolor{darkgreen}{clipper}, \textcolor{darkgreen}{government}, \textcolor{darkgreen}{chip}, \textcolor{darkgreen}{algorithm}
        & \textcolor{darkgreen}{encryption}, \textcolor{darkgreen}{clipper}, \textcolor{darkgreen}{chip}, \textcolor{darkgreen}{government}, \textcolor{darkgreen}{people}\\
        & \textcolor{mediumorchid}{faith}, \textcolor{mediumorchid}{christians}, \textcolor{mediumorchid}{church}, \textcolor{mediumorchid}{christ}, \textcolor{mediumorchid}{jesus}
        & \textcolor{mediumorchid}{church}, \textcolor{mediumorchid}{book}, \textcolor{mediumorchid}{christian}, \textcolor{mediumorchid}{life}, \textcolor{mediumorchid}{faith}
        & \textcolor{mediumorchid}{faith}, \textcolor{mediumorchid}{sell}, \textcolor{mediumorchid}{church}, \textcolor{mediumorchid}{bible}, \textcolor{mediumorchid}{condition}\\
        & \textcolor{orange}{turkish}, \textcolor{orange}{turkey}, \textcolor{orange}{turks}, \textcolor{orange}{armenians}, \textcolor{orange}{armenian}
        & \textcolor{orange}{soviet}, \textcolor{orange}{greek}, \textcolor{orange}{armenian}, \textcolor{orange}{turks}, \textcolor{orange}{turkish}\\
        &&\textcolor{red}{sale}, \textcolor{red}{sell}, \textcolor{red}{nntp}, \textcolor{red}{offer}, \textcolor{red}{host}
        & \textcolor{red}{sale}, \textcolor{red}{subject}, \textcolor{red}{line}, \textcolor{red}{card}, \textcolor{red}{offer}\\
        \midrule
        
        $t_{9}$ & \textcolor{darkblue}{software}, \textcolor{darkblue}{computer}, \textcolor{darkblue}{video}, \textcolor{darkblue}{system}, \textcolor{darkblue}{card}
        & \textcolor{darkblue}{disk}, \textcolor{darkblue}{game}, \textcolor{darkblue}{snes}, \textcolor{darkblue}{genesis}, \textcolor{darkblue}{games}
        & \textcolor{darkblue}{modem}, \textcolor{darkblue}{monitor}, \textcolor{darkblue}{disk}, \textcolor{darkblue}{card}, \textcolor{darkblue}{scsi}\\
        & \textcolor{darkgreen}{system}, \textcolor{darkgreen}{computer}, \textcolor{darkgreen}{software}, \textcolor{darkgreen}{access}, \textcolor{darkgreen}{phone}
        & \textcolor{darkgreen}{encryption}, \textcolor{darkgreen}{clipper}, \textcolor{darkgreen}{government}, \textcolor{darkgreen}{chip}, \textcolor{darkgreen}{algorithm}
        & \textcolor{darkgreen}{chip}, \textcolor{darkgreen}{clipper}, \textcolor{darkgreen}{encryption}, \textcolor{darkgreen}{escrow}, \textcolor{darkgreen}{phone}\\
        & \textcolor{mediumorchid}{christians}, \textcolor{mediumorchid}{life}, \textcolor{mediumorchid}{christian}, \textcolor{mediumorchid}{marriage}, \textcolor{mediumorchid}{jesus}
        & \textcolor{mediumorchid}{church}, \textcolor{mediumorchid}{book}, \textcolor{mediumorchid}{christian}, \textcolor{mediumorchid}{life}, \textcolor{mediumorchid}{faith}
        & \textcolor{mediumorchid}{faith}, \textcolor{mediumorchid}{jesus}, \textcolor{mediumorchid}{grace}, \textcolor{mediumorchid}{mary}, \textcolor{mediumorchid}{atheist}\\
        & \textcolor{orange}{jews}, \textcolor{orange}{jewish}, \textcolor{orange}{israel}, \textcolor{orange}{israeli}, \textcolor{orange}{arabs}
        & \textcolor{orange}{soviet}, \textcolor{orange}{greek}, \textcolor{orange}{armenian}, \textcolor{orange}{turks}, \textcolor{orange}{turkish}
        & \textcolor{orange}{write}, \textcolor{orange}{subject}, \textcolor{orange}{line}, \textcolor{orange}{israel}, \textcolor{orange}{article}\\
        &&\textcolor{red}{sale}, \textcolor{red}{sell}, \textcolor{red}{nntp}, \textcolor{red}{offer}, \textcolor{red}{host}
        & \textcolor{red}{sale}, \textcolor{red}{forsale}, \textcolor{red}{card}, \textcolor{red}{offer}, \textcolor{red}{subject}\\
        \midrule
        
        $t_{10}$ & \textcolor{darkblue}{software}, \textcolor{darkblue}{computer}, \textcolor{darkblue}{video}, \textcolor{darkblue}{system}, \textcolor{darkblue}{card}
        & \textcolor{darkblue}{disk}, \textcolor{darkblue}{game}, \textcolor{darkblue}{snes}, \textcolor{darkblue}{genesis}, \textcolor{darkblue}{games}
        & \textcolor{darkblue}{tempest}, \textcolor{darkblue}{edward}, \textcolor{darkblue}{reid}, \textcolor{darkblue}{scsi}, \textcolor{darkblue}{machine}\\
        & \textcolor{darkgreen}{system}, \textcolor{darkgreen}{computer}, \textcolor{darkgreen}{access}, \textcolor{darkgreen}{software}, \textcolor{darkgreen}{device}
        & \textcolor{darkgreen}{encryption}, \textcolor{darkgreen}{clipper}, \textcolor{darkgreen}{government}, \textcolor{darkgreen}{chip}, \textcolor{darkgreen}{algorithm}
        & \textcolor{darkgreen}{chip}, \textcolor{darkgreen}{encryption}, \textcolor{darkgreen}{clipper}, \textcolor{darkgreen}{line}, \textcolor{darkgreen}{subject}\\
        & \textcolor{mediumorchid}{christians}, \textcolor{mediumorchid}{life}, \textcolor{mediumorchid}{christian}, \textcolor{mediumorchid}{believe}, \textcolor{mediumorchid}{marriage}
        & \textcolor{mediumorchid}{church}, \textcolor{mediumorchid}{book}, \textcolor{mediumorchid}{christian}, \textcolor{mediumorchid}{life}, \textcolor{mediumorchid}{faith}
        & \textcolor{mediumorchid}{faith}, \textcolor{mediumorchid}{jesus}, \textcolor{mediumorchid}{grace}, \textcolor{mediumorchid}{religion}, \textcolor{mediumorchid}{atheist}\\
        & \textcolor{orange}{jews}, \textcolor{orange}{jewish}, \textcolor{orange}{israel}, \textcolor{orange}{muslim}, \textcolor{orange}{muslims}
        & \textcolor{orange}{soviet}, \textcolor{orange}{greek}, \textcolor{orange}{armenian}, \textcolor{orange}{turks}, \textcolor{orange}{turkish}
        & \textcolor{orange}{write}, \textcolor{orange}{israel}, \textcolor{orange}{article}, \textcolor{orange}{israeli}, \textcolor{orange}{arab}\\
        &&\textcolor{red}{sale}, \textcolor{red}{sell}, \textcolor{red}{nntp}, \textcolor{red}{offer}, \textcolor{red}{host}
        & \textcolor{red}{sale}, \textcolor{red}{forsale}, \textcolor{red}{card}, \textcolor{red}{offer}, \textcolor{red}{manual}\\
        \midrule
        
        $t_{11}$ & \textcolor{darkblue}{software}, \textcolor{darkblue}{computer}, \textcolor{darkblue}{video}, \textcolor{darkblue}{system}, \textcolor{darkblue}{card}
        & \textcolor{darkblue}{disk}, \textcolor{darkblue}{game}, \textcolor{darkblue}{snes}, \textcolor{darkblue}{genesis}, \textcolor{darkblue}{games}
        & \textcolor{darkblue}{monitor}, \textcolor{darkblue}{hardware}, \textcolor{darkblue}{scsi}, \textcolor{darkblue}{controller}, \textcolor{darkblue}{chip}\\
        & \textcolor{darkgreen}{system}, \textcolor{darkgreen}{computer}, \textcolor{darkgreen}{software}, \textcolor{darkgreen}{access}, \textcolor{darkgreen}{device}
        & \textcolor{darkgreen}{internet}, \textcolor{darkgreen}{anonymous}, \textcolor{darkgreen}{ripem}, \textcolor{darkgreen}{privacy}, \textcolor{darkgreen}{user}
        & \textcolor{darkgreen}{encryption}, \textcolor{darkgreen}{clipper}, \textcolor{darkgreen}{chip}, \textcolor{darkgreen}{public}, \textcolor{darkgreen}{subject}\\
        & \textcolor{mediumorchid}{life}, \textcolor{mediumorchid}{christians}, \textcolor{mediumorchid}{christian}, \textcolor{mediumorchid}{believe}, \textcolor{mediumorchid}{woman}
        & \textcolor{mediumorchid}{church}, \textcolor{mediumorchid}{book}, \textcolor{mediumorchid}{christian}, \textcolor{mediumorchid}{life}, \textcolor{mediumorchid}{faith}
        & \textcolor{mediumorchid}{religion}, \textcolor{mediumorchid}{mary}, \textcolor{mediumorchid}{faith}, \textcolor{mediumorchid}{jesus}, \textcolor{mediumorchid}{convert}\\
        & \textcolor{orange}{jews}, \textcolor{orange}{jewish}, \textcolor{orange}{muslims}, \textcolor{orange}{israel}, \textcolor{orange}{muslim}
        & \textcolor{orange}{soviet}, \textcolor{orange}{genocide}, \textcolor{orange}{armenian}, \textcolor{orange}{russian}, \textcolor{orange}{argic}
        & \textcolor{orange}{israel}, \textcolor{orange}{israeli}, \textcolor{orange}{arab}, \textcolor{orange}{write}, \textcolor{orange}{jewish}\\
        && \textcolor{red}{sale}, \textcolor{red}{host}, \textcolor{red}{nntp}, \textcolor{red}{sell}, \textcolor{red}{government}
        & \textcolor{red}{sale}, \textcolor{red}{forsale}, \textcolor{red}{card}, \textcolor{red}{offer}, \textcolor{red}{seagate}\\
    \bottomrule
    \end{tabular}
    }
    \label{tab:topics_extreme}
\end{table}

\end{document}